%% file: 00_Main.tex
\documentclass[10.5pt]{article}

\usepackage{relsize} 
\usepackage{amsmath} 
\usepackage{tabularx,booktabs} 
\usepackage{multirow} 
\usepackage{listings} 
\usepackage{subfiles} 
\usepackage{graphicx} 
\usepackage[hidelinks]{hyperref} 
\usepackage[yyyymmdd]{datetime} 
\usepackage{comment} 
\usepackage{enumitem} 
\setlist[enumerate]{label*=\arabic*.} 
\usepackage{bm} 
\usepackage{amssymb} 
\usepackage{orcidlink} 

\newcolumntype{C}{>{\centering\arraybackslash} X } 
\graphicspath{{figures/}} 

\newcommand{\clt}{%
    \mathchoice{\raisebox{1pt}{$\displaystyle <$}}
    {\raisebox{1pt}{$<$}}
    {\raisebox{0.8pt}{$\scriptstyle <$}}
    {\raisebox{0.2pt}{$\scriptscriptstyle <$}}}

\makeatletter
\AtBeginDocument{%
    \def\doi#1{\url{https://doi.org/#1}}}
\makeatother

\usepackage[showframe=false]{geometry} 
\usepackage{authblk}

\usepackage{fancyhdr}
\usepackage{indentfirst} 

\title{Machine learning for structural design models of continuous beam systems via influence zones}

\author[1,*]{\href{https://orcid.org/0000-0003-4939-9916}{\orcidlink{0000-0003-4939-9916} Adrien Gallet}}
\author[2]{\href{https://orcid.org/0000-0001-8229-9605}{\orcidlink{0000-0001-8229-9605} Andrew Liew}}
\author[1]{\href{https://orcid.org/0000-0003-2597-8200}{\orcidlink{0000-0003-2597-8200} Iman Hajirasouliha}}
\author[3]{\href{https://orcid.org/0000-0002-6730-5277}{\orcidlink{0000-0002-6730-5277} Danny Smyl}}

\affil[1]{Department of Civil and Structural Engineering, University of Sheffield, Sheffield UK}
\affil[2]{Unipart Construction Technologies, Rotherham, UK}
\affil[3]{School of Civil and Environmental Engineering, Georgia Institute of Technology, Atlanta, GA, USA}
\affil[ ]{ } 
\affil[*]{\normalsize Author for correspondence: \textit{agallet1@sheffield.ac.uk}}
\affil[ ]{ } 
\affil[ ]{\normalsize Version: Accepted Manuscript (AM)}

\date{Date: 2024.03.12} 

\begin{document}
    
    \maketitle 
    \thispagestyle{empty} 
    
    \begin{abstract}
        \noindent This work develops a machine learned structural design model for continuous beam systems from the inverse problem perspective. After demarcating between forward, optimisation and inverse machine learned operators, the investigation proposes a novel methodology based on the recently developed influence zone concept \emph{which represents a fundamental shift in approach compared to traditional structural design methods}. The aim of this approach is to conceptualise a non-iterative structural design model that predicts cross-section requirements for continuous beam systems of arbitrary system size. After generating a dataset of known solutions, an appropriate neural network architecture is identified, trained, and tested against unseen data. The results show a mean absolute percentage testing error of 1.6\% for cross-section property predictions, along with a good ability of the neural network to generalise well to structural systems of variable size. The CBeamXP dataset generated in this work and an associated python-based neural network training script are available at an open-source data repository to allow for the reproducibility of results and to encourage further investigations.
    \end{abstract}
    

    \subfile{01_Introduction}
    \subfile{02_Problem_statement}

    \subfile{03_Methodology}
    \subfile{04_Results}
    \subfile{05_Discussion}
    \subfile{06_Conclusion}
    \subfile{07_Acknowledgements}
    
    \bibliographystyle{unsrt-numeric-style}
    \bibliography{08_References} 
    
\end{document}

%% file: 01_Introduction.tex
\clearpage

\section{Introduction}
\label{sec:introduction}

\begin{figure*}[htb]
    \centering
    \includegraphics[width=\textwidth,height=\textheight,keepaspectratio]{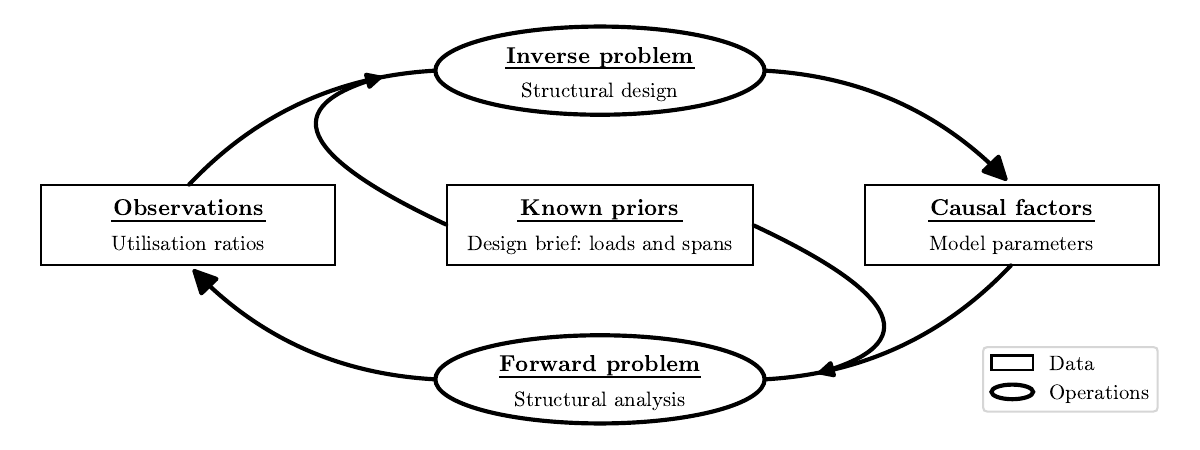}
    \caption{The inverse problem perspective for structural design, which relies on known priors such as design brief details of loading and span requirements along with observations of utilisation ratios that represent structural adequacy to evaluate the model parameters of a solution, such as size, shape and topology of a viable structure. Structural analysis is treated as the forward problem.}
    \label{fig:inverse_problem_perspective}
\end{figure*}

It was recently argued that structural design is an \textit{inverse problem} \cite{GalletA_etal_2022_Structural_engineering_from}, in which one estimates the \textit{model parameters} (the causal factors) of possible structural solutions from a set of \textit{structural utilisations} (the observations). This inverse problem perspective, highlighted in Figure \ref{fig:inverse_problem_perspective}, is
underscored by the ill-posed characteristics structural design shares with other inverse problems \cite{AlifanovO_1983_Methods_of_solving}, which in civil and structural engineering include subject areas such as structural health monitoring \cite{FriswellM_2007_Damage_identification_using, SmylD_etal_2018_Detection_and_reconstruction}, self-sensing smart materials \cite{TallmanT_etal_2020_Structural_health_and, ZhaoL_etal_2021_Spatial_Damage_Characterization} and forensic blast engineering \cite{RigbyS_etal_2020_Preliminary_yield_estimation, vanderVoortM_etal_2015_Forensic_analysis_of}.

Inverse problems are predominantly solved iteratively \cite{MuellerJ_etal_2012_Linear_and_Nonlinear}, and unsurprisingly so is structural design \cite{MasonJ_etal_2011_Structural_design__the}, often with the help of structural optimisation such as size \cite{PolliniN_2021_Gradient_based_prestress_and, AltayO_etal_2023_Size_optimization_of}, shape \cite{DaxiniS_etal_2017_Parametric_shape_optimization, UpadhyayB_etal_2021_Numerical_analysis_perspective}, topology \cite{HajirasoulihaI_etal_2011_Topology_optimization_for, SigmundO_etal_2013_Topology_optimization_approaches_} and layout optimisation \cite{WeldeyesusA_etal_2019_Adaptive_solution_of, FaircloughH_etal_2022_Layout_optimization_of}. Provided that a clear objective function exists, these techniques are the state of the art for solving the structural design inverse problem \textit{iteratively}.

However, in industry, the uptake of iteration based design approaches face certain barriers, including high computational costs \cite{ZaheerQ_etal_2023_A_review_on}, complex outputs that require additional post-rationalisation \cite{HeL_etal_2015_Rationalization_of_trusses}, and demand a particular expertise from practising design engineers that can be absent from engineering curriculums \cite{CoelhoR_etal_2014_Form_finding_&}. These challenges have encouraged researchers to investigate the use of machine learning (ML) methodologies for structural design \cite{Malaga-ChuquitaypeC_2022_Machine_Learning_in}. This parallels a similar development of using ML within the domain of inverse problems \cite{ArridgeS_etal_2019_Solving_inverse_problems}, with exemplary applications in areas such as structural health monitoring \cite{ChenL_etal_2022_Probabilistic_cracking_prediction, HamiltonS_etal_2018_Deep_D_Bar__Real_Time}, that aid or replace the optimisation problem with \textit{learned} components.

The earliest application of such machine learned components for structural design occurred in 1989 \cite{AdeliH_etal_1989_Perceptron_Learning_in} with simplified perceptron models. This research was followed in the 1990s by more advanced feed-forward neural networks for simple reinforced concrete beam depth estimations \cite{VanlucheneR_etal_1990_Neural_Networks_in} as well as cross-sectional area predictions of trusses \cite{BerkeL_etal_1993_Optimum_design_of, KangH_etal_1994_Neural_Network_Approaches}. Whilst other machine learning modalities such as support vector machines \cite{HannaS_2007_Inductive_machine_learning} have also been studied, neural networks tend to outperform other ML models archetypes in terms of prediction error \cite{TseranidisS_etal_2016_Data_driven_approximation_algorithms}.

More recently, deep learning techniques have been investigated for structural design. These include convolutional \cite{BehzadiM_etal_2021_Real_Time_Topology_Optimization, XiangC_etal_2022_Accelerated_topology_optimization} and generative adversarial networks \cite{NieZ_etal_2021_TopologyGAN__Topology_Optimization, LiaoW_etal_2021_Automated_structural_design} to accelerate topology optimisation, and the application of variational auto encoders for structural design space exploration \cite{DanhaiveR_etal_2021_Design_subspace_learning_}. A common limitation across such investigations is the inability for the same machine learned model to generalise to differently sized topologies and structural arrangements. These two challenges, highlighted by design ill-posedness and the inability of previous machine learning models to generalise to structural arrangements of arbitrary size, have motivated the work presented here.

This investigation has two objectives. The first objective is to reconcile the relationship between structural design, inverse problems and machine learning by developing a non-iterative structural design model for continuous beam systems using a multi-layer neural network. The authors believe that this perspective could serve as a framework to distinguish between different types of machine learning applications within the field of structural engineering in the future. The second objective is to address the inherent issue of generalisability in respect to system size by taking advantage of a recently developed concept known as a continuous beam's influence zone \cite{GalletA_etal_2023_Influence_zones_for}. This technique could potentially form the basis to generalise a design model for continuous structural systems of arbitrary topology, and might complement other techniques that attempt to address the generalisability issue such as graph neural networks \cite{WhalenE_etal_2022_Toward_Reusable_Surrogate, BlekerL_etal_2023_Structural_Form_Finding_Enhanced}.

The paper is structured as follows: Section \ref{sec:problem_statement} explores the problem statement from the inverse problem perspective and provides the rationale for machine learned design models, Section \ref{sec:methodology} explains the methodology employed to develop the generalisable structural design model, Section \ref{sec:results} presents the step-by-step process of the neural network development process, and Section \ref{sec:discussion} discusses the model's generalisability and prediction variability, along with suggestions for further research.

%% file: 02_Problem_statement.tex
\section{Problem statement}
\label{sec:problem_statement}
\subsection{A novel perspective}

The inverse problem perspective for structural design as shown in Figure \ref{fig:inverse_problem_perspective} consists out of two operations, the \textit{forward} and \textit{inverse problem} (shown as the bottom and top ellipses, respectively) and three sets of data: \textit{observations}, \textit{known priors} and \textit{causal factors} (shown as rectangles from left to right, respectively). One of the underpinning features of the inverse problem perspective is the clear demarcation between structural analysis and structural design, a distinction often re-iterated in engineering philosophy \cite{KoenB_2003_Discussion_of_the, BulleitW_etal_2015_Philosophy_of_Engineering_}, yet never linked to the corresponding nature of forward and inverse problems, respectively.

Both the forward (structural analysis) and the inverse problem (structural design) rely on \textit{known priors}, shown centrally in Figure \ref{fig:inverse_problem_perspective}, which can be thought of as constraints set by a design brief such as load and span requirements. During design they inform and regularise the search space of \textit{causal factors} (model parameters such as section properties and topologies), and in analysis they allow the evaluation of \textit{observations} (utilisation ratios such as ultimate (ULS) and serviceability limit states (SLS) \cite{CEN_2002_BS_EN_1990_2002_A1_2005}). Unlike traditional inverse problems, the observations are not measured physically, yet are expressed theoretically based on the utilisation ratios that could be measured from a compliant design solution which the set of causal factors correspond with; inverse problems are not defined by the physicality of the observations.

Within this context, the application of machine learning in structural engineering can be split into three categories based on the type of operations the machine learned components replace. These categories help distinguish between fundamentally different types of machine learning applications that occur within the context of structural engineering and are identifiable across different decades of the literature:

\begin{enumerate}[label=\alph*)]
    \item \textit{ML forward operators}: machine learned components that aid or accelerate solving the forward problem (structural analysis) to inform or validate design decisions. Examples include neural network like models as quick re-analysis tools for optimum design (1991) \cite{HajelaP_etal_1991_Neurobiological_computational_models} and machine learning models to determine the buckling behaviour and model decomposition of thin-walled members required for structural analysis (2023) \cite{MojtabaeiS_etal_2023_Predicting_the_buckling}.
    \item \textit{ML optimisation solvers}: machine learned components entirely motivated by the traditional iterative solution process to arrive at structural designs. Examples include ``neural dynamic models'' developed as an alternative structural design optimisation technique (1995) \cite{AdeliH_etal_1995_Optimization_of_space} and a physics informed neural energy-force network that replaces both the structural design and analysis steps (2023) \cite{MaiH_etal_2023_Physics_informed_neural_energy_force}.
    \item \textit{ML inverse operators}: machine learned components which solve the inverse problem (structural design) by mapping a set of structural utilisations and known priors to model parameters directly. Examples include estimating cross-sectional properties for simple trusses directly based on known optimum examples using neural networks (1994) \cite{KangH_etal_1994_Neural_Network_Approaches} and approximating topological optimised structures in real-time using convolutional neural networks (2022) \cite{YanJ_etal_2022_Deep_learning_driven}.
\end{enumerate}

\begin{figure*}[htb]
    \centering
    \includegraphics[width=\textwidth,height=\textheight,keepaspectratio]{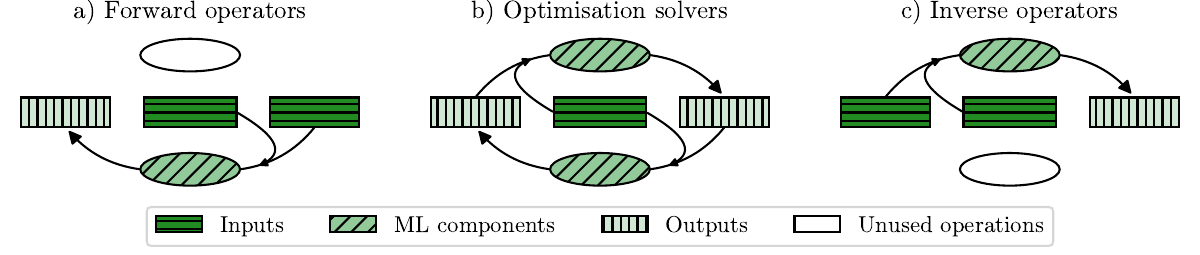}
    \caption{Types of machine learning (ML) components from the inverse problem perspective. Shapes in each sub-figure; ellipses: inverse problem (top), forward problem (bottom); rectangles: observations (left), known priors (middle), causal factors (right).}
    \label{fig:ML_components_for_design}
\end{figure*}

These three categories can also be differentiated visually as shown in Figure \ref{fig:ML_components_for_design}. It is worth noting that the field of ML forward operators has likely received the most research attention in the form of ``surrogate models'' \cite{WangG_etal_2007_Review_of_Metamodeling_a, KozielS_etal_2013_Surrogate_Based_Modeling_and}. In this respect, machine learned optimisation solvers and inverse operators are less common. Furthermore, the machine learned forward operators and optimisation solvers identified above typically require some form of iteration to achieve structural design; machine learned inverse operators on the other hand can be non-iterative \cite{KangH_etal_1994_Neural_Network_Approaches, YanJ_etal_2022_Deep_learning_driven}. The ability to provide real-time design feedback is of particular interest to address the limitations of current iterative structural design approaches. To this end, and in support of the inverse problem perspective, this paper will focus on developing a non-iterative structural design model for continuous beam systems.

\subsection{Design problem: continuous beam systems}
\label{sec:problem_statement_detailed}
Continuous beam systems arise in structural engineering when rigid connections between members are required or unavoidable due to design or material considerations. The support fixity and structural connectivity render the system statically indeterminate. This poses a challenge from a design perspective, since the compliance of cross-sectional properties cannot be evaluated without knowledge of their magnitudes; this results in an iterative design process, especially for complex design scenarios with heterogeneous loading and span conditions \cite{SakaM_etal_2013_Mathematical_and_Metaheuristic}.

\begin{figure}[tbh]
    \centering
    \includegraphics[width=\textwidth,height=\textheight,keepaspectratio]{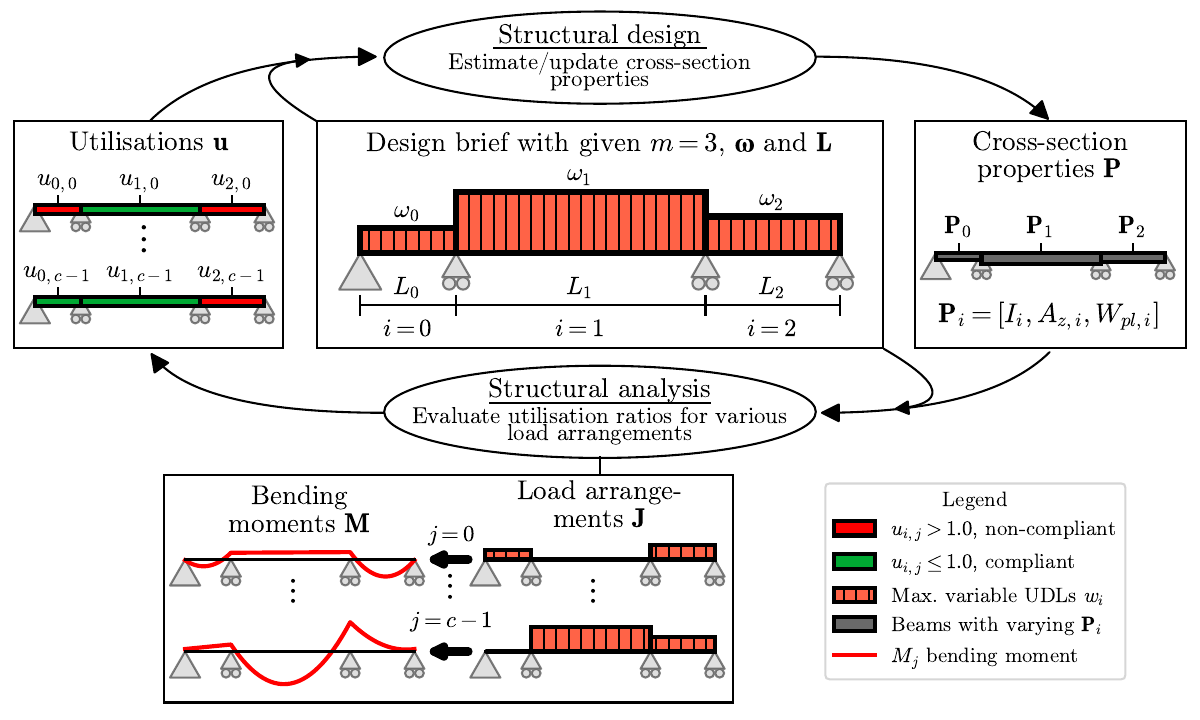}
    \caption{Design process of a continuous beam system from the inverse problem perspective.}
    \label{fig:design_beam_system}
\end{figure}

Figure \ref{fig:design_beam_system} highlights the design problem for continuous beam systems from the inverse problem perspective. The known priors, which are shown centrally as the design brief, include the number of members $m$ in the system indexed by $i$ with span length $L_i$ from vector $\mathbf{L} = [L_i]_{0\le i \clt m}$, subjected to uniformly distributed loads (UDLs) $\omega_i$ from vector $\bm{\omega} = [\omega_i]_{0\le i \clt m}$. These known priors and the utilisation ratios $\mathbf{u}$ of the members, shown on the left in Figure \ref{fig:design_beam_system}, are needed to evaluate the causal factors, shown on the right as the cross-section property vector $\mathbf{P} = [\mathbf{P}_i]_{0\le i \clt m}$.

The design problem is complicated due to the existence of $c$ potentially critical load arrangements $J$ indexed by $j$ from set $\mathbf{J} = [J_j]_{0\le j \clt c}$ shown at the bottom of Figure \ref{fig:design_beam_system}. The size $c$ of $\mathbf{J}$ was studied in \cite{GalletA_etal_2023_Influence_zones_for}. Each of these load arrangements cause different structural responses such as bending moments $\mathbf{M}$, and will give rise to a matrix of utilisation ratios $u_{i,j}$ to form matrix $\mathbf{u} = [u_{ij}]_{0\le i \clt m,\, 0\le j \clt c}$ that can be evaluated with structural analysis to check for structural compliance ($u_{i,j} \le 1.0$). Instead of repeatedly assuming cross-section properties $\mathbf{P}$ and conducting structural analysis calculations until the matrix of utilisation ratios $\mathbf{u}$ are compliant, a machine learned inverse operator relies solely on the known priors and the utilisation ratios $\mathbf{u}$ to directly evaluate the cross-section properties $\mathbf{P}$.

For the purpose of the continuous beam system considered in this work, several assumptions will be made: members are made out of S355 steel, are considered laterally restrained (and hence not susceptible to lateral instability), Timoshenko-Ehrenfest beam theory is used to model this system and the structure will be analysed elastically yet designed against ULS plastic cross-section property checks as allowed by Eurocode EN 1993-1-1 5.4.2 (2) \cite{CEN_2015_BS_EN_1993_1_1_2005_A1_2014}. The cross-sectional properties to be evaluated include the major axis second moment of area $I$, the major axis shear area $A_{z}$ and the major axis plastic section modulus $W_{\mathrm{pl}}$ for each member $i$. Together they form the member-based cross-sectional property vector $\mathbf{P}_i$:

\begin{equation}
    \label{eq:cross_section_vector}
    \mathbf{P_i} = [I_{i}, A_{z,i}, W_{\mathrm{pl},i}]
\end{equation}

\noindent The structural analysis operation in Figure \ref{fig:design_beam_system} is defined by a forward operator:

\begin{equation}
    \begin{gathered}
        \label{eq:forward_operator_general}
        \mathbf{u} = O_{\mathrm{forw}}(m,\bm{\omega},\mathbf{L},\mathbf{P})
    \end{gathered}
\end{equation}

\noindent and similarly the structural design operation by an inverse operator:

\begin{equation}
    \begin{gathered}
        \label{eq:inverse_operator_general}
        O_{\mathrm{inv}}(m,\bm{\omega},\mathbf{L},\mathbf{u}) = \mathbf{P}
    \end{gathered}
\end{equation}

\noindent where both $O_{\mathrm{forw}}$ and $O_\mathrm{inv}$ rely on the same known priors, the design brief information $m$, $\bm{\omega}$ and $\mathbf{L}$ that define the structural system and design problem.

\subsection{The need for machine learned inverse operators}

Defining an explicit non-iterative inverse operator for Equation \ref{eq:inverse_operator_general} is challenging due to the difficulty of inverting the forward operator and is directly linked to the ill-posed nature common across most inverse problems \cite{AlifanovO_1983_Methods_of_solving}. A quantitative evaluation of the extent of ill-posedness in structural design is not obvious, however it is possible to describe why the structural design problem shown in Figure \ref{fig:design_beam_system} is ill-posed, namely due to the infinite number of viable solutions, and:

\begin{enumerate}[label=\alph*)]
    \item The physical limitations introduced by yielding, buckling, serviceability that arise when combining the forward model with structural codes, resulting in a discontinuous relationship between the observations and causal factors.
    \item The indeterminacy of the continuous beam system which increases with the number of members of the system.
\end{enumerate}

Furthermore, any structural analysis forward operators themselves are approximations of the true behaviour of structures, and dealing with this associated uncertainty is a key challenge in design. For example, engineers need to decide if the assumptions and simplifications of structural analysis models, such as the material response (e.g. perfectly elastic) and underlying beam theory (e.g. Euler–Bernoulli theory), are representative of the structure's true behaviour.

The difficulty of inverting a forward operator can be shown mathematically. Typically, the $O_{\mathrm{forw}}$ operator contains two steps. The first step, defined by $O_{\mathrm{forw},1}$ would evaluate the structural response of the system when subjected to a set of external forces $\bm{\omega}$ in terms of deflections and internal forces, and the second step, defined by $O_{\mathrm{forw},2}$, would take these structural response observations to evaluate the utilisation ratios based on design codes. Consider for example building a $O_{\mathrm{forw},1}$ operator using the stiffness matrix method to evaluate the internal force vector $[f_{p}]_i$ for member $i$ defined as:

\begin{equation}
    [f_{p}]_i = [V_{1,i}, M_{1,i}, V_{2,i}, M_{2,i}]^\intercal
\end{equation}

\noindent where $V$, and $M$ represent the internal shear forces and bending moments within member $i$ at the start (index $1$) and end of the member (index $2$). Let us also assume, for simplicity, that the members consist out of steel with $E$ and $G$ for the Youngs and shear modulus, respectively, with a maximum yield stress of $\sigma_y$. In this case, the internal forces $[f_{p}]_i$ for each member could be evaluated using a simplified Timoshenko-Ehrenfest beam theory for a \textit{single} load arrangement by Equation \ref{eq:internal_force_vector}. To achieve this, $[k_{pq}]_i$ is defined as the local stiffness matrix shown in Equation \ref{eq:local_stiffness_matrix}, $[K_{pq}]$ as the global stiffness matrix in Equation \ref{eq:global_stiffness_matrix}, $[d_q]$ as the nodal displacement vector in Equation \ref{eq:nodal_displacement_vector} with $F_p([\omega_i])$ as the external force vector, where rows and columns of all matrices are indexed by $p$ and $q$, respectively:

\begin{equation}
    \label{eq:internal_force_vector}
    [f_{p}]_i = [k_{pq}]_i [d_q]
\end{equation}

\begin{equation}
    \label{eq:local_stiffness_matrix}
    [k_{pq}]_i = \frac{EI_i}{L_i^3 (1-\varphi)}
    \begin{bmatrix}
        12 & 6L_i & -12 & 6L_i \\
        6L_i & 4L_i^2 & -6L_i & 2L_i^2 \\
        -12 & -6L_i & 12 & -6L_i \\
        6L_i & 2L_i^2 & -6L_i & 4L_i^2
    \end{bmatrix}
    , \quad \varphi = \frac{12EI_i}{A_z G L_i^2}
\end{equation}

\begin{equation}
    \label{eq:global_stiffness_matrix}
    [K_{pq}] = [k_{pq}]_0 + [k_{pq}]_1 + \ldots + [k_{pq}]_{m-1} = \sum_{i=0}^{m-1} [k_{pq}]_i
\end{equation}

\begin{equation}
    \label{eq:nodal_displacement_vector}
    [d_q] = [K_{pq}]^{-1} \, [F_p([\omega_i])]
\end{equation}

These operations can be succinctly written to transform the cross-section vector $\mathbf{P}$ with help of the known priors $m$, $\bm{\omega}$, $\mathbf{L}$ into the internal forces vector for each member $i$:

\begin{equation}
    O_{\mathrm{forw}, 1, i} = [f_{p}]_i =  [k_{pq}]_i  {\left[ \sum_{i=0}^{m-1} [k_{pq}]_i \right]}^{-1} \Biggr[ F_p([\omega_i]) \Biggr]
\end{equation}

Inverting this equation to yield $O_{\mathrm{inv}}$ is difficult since it would require separating or decomposing the individual cross-section properties $\mathbf{P}_i$ out of the stiffness matrices $[k_{pq}]_i$. This cannot be done without, at minimum, making some assumptions about the relative proportions of the cross-section properties from one member to another. Inverting the second step of the forward operator $O_{\mathrm{forw}, 2}$ poses further challenges. Suppose $O_{\mathrm{forw}, 2}$ transforms the internal member forces $[f_{p}]_i$ to evaluate the governing (critical) utilisation ratios indexed by $r$ for $t$ design checks for a single load arrangement $J$. For example, using the steel design code EN 1993-1-1 \cite{CEN_2015_BS_EN_1993_1_1_2005_A1_2014}:

\begin{equation}
    \begin{split}
        u_i = O_{\mathrm{forw}, 2, i}(O_{\mathrm{forw}, 1, i}&) = \max([u_{ir}]_{0\le i \clt m,\, 0\le r \clt t}) = \max(u_{i,0}, u_{i,1}, \cdots, u_{i,t-1}) \\
        \text{where:} \\
        u_{i,0} &= \frac{V_{1,i}}{A_{z,i} \, \sigma_y / \sqrt{3}}, \quad
        u_{i,1} = \frac{M_{1,i}}{W_{\mathrm{pl},i} \, \sigma_y} \\
        u_{i,2} &= \frac{V_{2,i}}{A_{z,i} \, \sigma_y / \sqrt{3}}, \quad
        u_{i,3} = \frac{M_{2,i}}{W_{\mathrm{pl},i} \, \sigma_y} \\
        & \quad \quad \quad \quad \quad \quad \quad \quad \vdots \\
        &u_{i,t-1} \text{ for other compliance checks}
    \end{split}
\end{equation}

The difficulty here is that the governing utilisation ratio can change according to the known priors of the problem statement. This means that an individual equation for each possible critical design check would need to be derived. For example different design equations exist for the same structural check depending on the type of cross-section (Class 1 vs. Class 4) \cite{CEN_2015_BS_EN_1993_1_1_2005_A1_2014} a final design solution might contain, which is not known ahead of time. Note also that the equations above do not even consider the serviceability limit state, the multiple load arrangements $\mathbf{J}$ which may be critical, nor the need to sufficiently discretise individual beam members.

It is because of the challenges identified above that machine learned inverse operators are particularly appealing, since they can approximate a relationship between a set of variables that may be difficult to encode explicitly \cite{MitchellT_1997_Machine_Learning}. Given a dataset generated by the $O_{\mathrm{forw}}$ operator that maps a set of cross-section properties $\mathbf{P}$ to compliant utilisations ratios $\mathbf{u}$, one can train a probabilistic machine learning model $O^{\dagger}_{\mathrm{inv}}$ with parameters $\bm{\theta}$ to map the set of bounded utilisation ratios $\mathbf{u}$ back to the cross-sectional properties $\mathbf{P}$ with known priors $m$, $\bm{\omega}$ and $\mathbf{L}$:

\begin{equation}
    \label{eq:ML_inverse_operator}
    O^{\dagger}_{\mathrm{inv}}(\bm{\theta}, m, \bm{\omega}, \mathbf{L}, \mathbf{u}) \approx \mathbf{P}
\end{equation}

By generating a dataset of valid structural designs with the help of existing optimisation approaches that contain the forward operator, a supervised machine learning model can be trained to learn the mapping of known priors and utilisations to cross-sectional properties directly. This represents a fundamental shift from traditional approaches employed in structural design that rely on engineering expertise and computationally expensive structural analysis or optimisation models at the point of design application. Machine learned inverse operators create non-iterative structural design models for which there currently exist no explicitly defined equivalents. Instead of focusing on accelerating forward models, computational resources can be invested in generating a dataset using physically complex yet realistic modelling assumptions. These machine learned structural design models aim to provide significantly greater generalisability than typical rules of thumb employed in design whilst still providing real-time feedback, benefit non-expert stakeholders whose own decision making relies on structural design outcomes and improves design knowledge permanence which can be difficult to attain due to industry turnover.

%% file: 03_Methodology.tex
\section{Methodology}
\label{sec:methodology}
\subsection{Choosing an appropriate machine learning model archetype}
\label{sec:method_ML_archetype}
The aim of the inverse operator $O^{\dagger}_{\mathrm{inv}}$ is to predict the cross-section property vector defined by Equation \ref{eq:cross_section_vector} numerically; therefore $O^{\dagger}_{\mathrm{inv}}$ will be a regression model. This restricts the types of supervised machine learning models of interest. The complexity and size of the design space are likely to demand a large dataset size discouraging the use of instance-based models such as the \textit{k}-nearest neighbour algorithm that store similarity measurements in memory \cite{KubatM_2021_An_introduction_to}. Similarly, support and relevance vector machines become impractical for datasets containing more than 3000 samples \cite{SmolaA_etal_2004_A_tutorial_on}. The non-linearity of the design problem voids the applicability of linear regression models, and decision trees (including the ensembled variants such as random forests) perform better at classification tasks \cite{BreimanL_2001_Random_Forests}.

These reasons motivated the use of neural networks, in particular multilayer neural networks (MLPs), a choice which is supported by evidence that suggests neural networks outperform other data-driven approximation algorithms in structural engineering applications \cite{TseranidisS_etal_2016_Data_driven_approximation_algorithms}. Although various archetypes exist ranging from convolutional (CNNs), recurrent (RNNs) and graph-based types (GNNs), MLPs are commonly used in literature \cite{BerkeL_etal_1993_Optimum_design_of, KangH_etal_1994_Neural_Network_Approaches}, and the results within this work could prove useful as a comparative performance measure for more advanced deep learning architectures \cite{BehzadiM_etal_2021_Real_Time_Topology_Optimization, NieZ_etal_2021_TopologyGAN__Topology_Optimization, DanhaiveR_etal_2021_Design_subspace_learning_} in future studies.

Multilayer neural networks have a fixed-dimensional input vector $\mathbf{x}_0$ of size $n$ that map to the output vector $\mathbf{x}_D$ of size $o$ with $D$ layers. In this study, a network has $D-1$ hidden layers of height $H$ each indexed by $d$ and are defined by $f_d(\mathbf{x}_d)$, which contains a (non-linear) activation function $a_d$ with weight matrix $\mathbf{w}_d$ and bias vector $\mathbf{b}_d$. The weight matrices and bias vectors of each layer form the model's parameters $\bm{\theta}=(\mathbf{w}=[\mathbf{w}_d]_{0\le d \clt D}, \mathbf{b}=[\mathbf{b}_d]_{0 \le d \clt D})$. Multiple hidden layers give form to the neural network $f$ through a function composition defined as:


\begin{equation}
    \begin{gathered}
        \label{eq:MLNN_equation}
        f: \mathbb{R}^n \rightarrow \mathbb{R}^o \\
        f(\mathbf{x}_0) \rightarrow \mathbf{x}_D: f_{D-1} \circ \ldots \circ f_1 \circ f_0(\mathbf{x}_0) \\
        f_{d}(\mathbf{x}_d) = a_d(\mathbf{w}_d \mathbf{x}_d + b_d) \\
    \end{gathered}
\end{equation}

The exact choice of architecture in terms of depth $D$, height $H$ and activation functions $a_d$ of the network as indicated in Equation \ref{eq:MLNN_equation} will require experimentation to achieve acceptable performance with a good bias-variance trade-off \cite{MurphyK_2022_Probabilistic_machine_learning_}. More importantly though, the features used for the input vector $\mathbf{x}_0$ will require careful consideration to create a generalisable inverse operator $O^{\dagger}_{\mathrm{inv}}$ as set out in Equation \ref{eq:ML_inverse_operator}.

\subsection{Selecting appropriate neural network features}
\label{sec:method_NN_features}
Feature selection, the process of choosing appropriate inputs, is essential for a machine learning model to generalise well to unseen data points. Unnecessary or irrelevant features can cause a model to learn a relationship with target variables that are not representative of the physical behaviour of the system, and thereby lead to worse results when interpolating within or extrapolating beyond the training set.

Previous studies of neural network based design models selected features relevant to the singular topology of the structural system at hand \cite{KangH_etal_1994_Neural_Network_Approaches, TseranidisS_etal_2016_Data_driven_approximation_algorithms}. Such approaches expose the largest limitation of multilayer neural networks: the fixed-dimensionality of the input vector \cite{WhalenE_etal_2022_Toward_Reusable_Surrogate}. These models may perform well for the particular topology they were trained against, yet the same model tends to perform worse or may not be applicable for differently sized structural systems, which severely limits their utility.

To address this limitation, this work takes advantage of a recently developed concept known as the \textit{influence zone} \cite{GalletA_etal_2023_Influence_zones_for}. The influence zone $k_\mathrm{max}$ is a measure of the extent to which surrounding design information is relevant for the utilisation evaluation of members. Whilst $k_\mathrm{max}$ differs for each member within a continuous structural system as shown in Figure \ref{fig:influence_zone_example}, for well defined design constraints and error thresholds, the maximum value of $k_\mathrm{max}$ within continuous beam systems converges towards a non-negative integer. The influence zone of member $g$ is found when the following two conditions are met:

\begin{figure}[htb]
    \centering
    \includegraphics[width=\textwidth,height=\textheight,keepaspectratio]{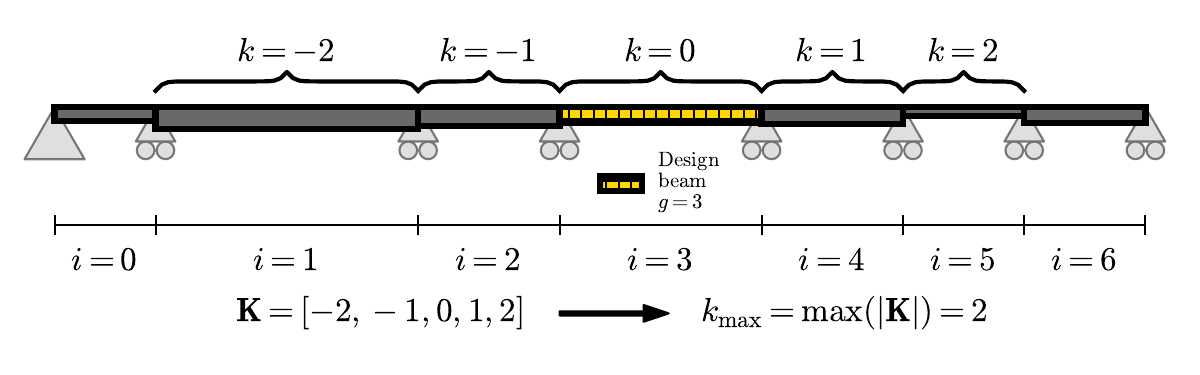}
    \caption{A figurative influence zone of $k_\mathrm{max}=2$ for design beam $g=3$ within a $m=7\textbf{}$ continuous beam system with $\epsilon_{\mathrm{max}} = 0.02$ limit.}
    \label{fig:influence_zone_example}
\end{figure}

\begin{equation}
    \begin{gathered}
        \label{eq:global_formulation}
        \left| \, 1 - \frac{u_{g,\mathrm{cap}}}{u_{g,\mathrm{true}}} \, \right| \le \epsilon_\mathrm{max} \\[0.4cm]
        u_{g,\mathrm{cap}} = \max \left( \ \mathlarger{\sum_{i=-k_\mathrm{max}}^{k_\mathrm{max}}} \mathbf{u}_{g,i,\,j}(\bm{\omega}, \mathbf{L}, \mathbf{P}, \mathbf{J}) \ \right)
    \end{gathered}
\end{equation}

In Equation \ref{eq:global_formulation}, $\epsilon_{\mathrm{max}}$ represents the maximum error threshold due to the difference between $u_{g,\mathrm{cap}}$, the captured utilisation ratio of the design beam $g$ for a given value of $k_\mathrm{max}$, and $u_{g,\mathrm{true}}$, the true utilisation ratio of the design beam $g$ if the contribution of all members of the continuous beam system had been considered. $\mathbf{u}_{g,i,\,j}$ is the utilisation ratio contribution function towards the design beam $g$ by member $i$ based on the UDLs $\bm{\omega}$, spans $\mathbf{L}$, structural properties $\mathbf{P}$ and load arrangements $\mathbf{J}$. If the requirement for $\epsilon_{\mathrm{max}}$ is sufficiently relaxed, the maximum influence zone $k_\mathrm{max}$ can be determined for any potential continuous beam system arising under the specified design constraints \cite{GalletA_etal_2023_Influence_zones_for}. This is extremely useful to ensure the relevant inputs are fed to a machine learning model. The influence zone thereby acts as a mechanics-driven feature selection process, and provides the basis to generalise to a continuous beam system of arbitrary size $m$.

\subsection{Structuring features for arbitrary system size \texorpdfstring{$m$}{\textit{m}}}
\label{sec:structuring_features_for_arbitrary_m}
\begin{figure}[htb]
    \centering
    \includegraphics[width=\textwidth,height=\textheight,keepaspectratio]{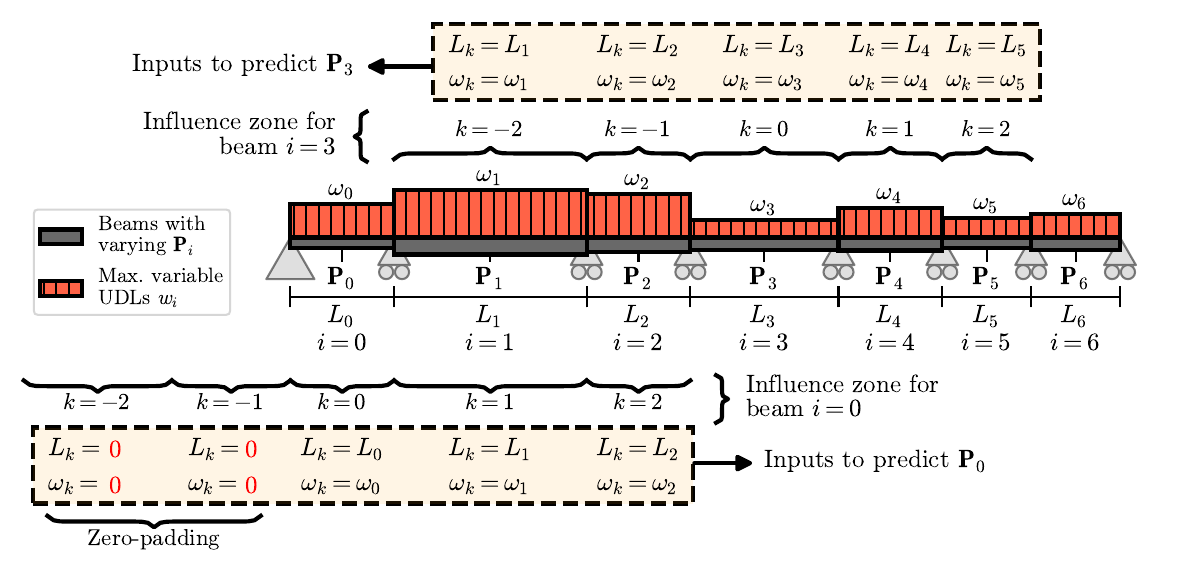}
    \caption{An illustration demonstrating the structuring of the neural network inputs using influence zones and zero-padding with $k_\mathrm{max} = 2$, leading to $n=4 k_\mathrm{max} + 2 = 10$ inputs.}
    \label{fig:zero_paddding_inputs}
\end{figure}

Zero-padding, the process of adding zero-valued inputs, arises in the context of convolutional neural networks to allow trained kernel filters to parse through the edges and corners of an input space \cite{MurphyK_2022_Probabilistic_machine_learning_}. This technique can also be applied to continuous beam systems to conceptualise a design model that parses over a structural system to make localised predictions for each member $i$. If the design information, here the UDLs $\bm{\omega}$ and span $\mathbf{L}$ that fall within the influence zone are provided as inputs to the network, then this would result in an input vector $\mathbf{x}_0$ of size $n = 4 k_\mathrm{max}+2$, as shown in Figure \ref{fig:zero_paddding_inputs} for member $i=3$ and $i=0$. These inputs should, based on the principle of the influence zone, contain the relevant information to predict the cross-section properties of member $d$ with an accuracy of up to $\epsilon_{\mathrm{max}}$.

It is now conceivable that the same neural network could be used to make a prediction for any other member using a fixed-dimensional input vector $\mathbf{x}_0$ by structuring the inputs relative to the position of the design beam's influence zone. This would include end-span beams by using zero-padding as shown in Figure \ref{fig:zero_paddding_inputs} for member $i=0$. Zero-padding in this instance is also logically consistent, since it corresponds with a beam that does not in fact exist; that is a beam of zero length $L$ and zero UDL load $\omega$. Therefore, instead of structuring the neural network based on the absolute position of a beam within the entire continuous beam system (as indexed by $i$), the inputs are structured relatively to the influence zone of a design beam $g$ to predict the cross-section properties of that design beam $\mathbf{P}_g$.

Whilst such an approach will require $m$ forward passes (inferences) to predict the cross-section properties of an $m$ sized system (one prediction per beam), it enables the same neural network to be applied to continuous beam systems of any size $m$ for which the maximum influence zone value $k_\mathrm{max}$ that determined the size of the input vector $\mathbf{x}_0$ applies to. Based on the principle of influence zones, the neural network will be able to make predictions for continuous beam systems of size greater than the fixed-dimensional input vector size $m>2 k_\mathrm{max} + 1$, since any information outside the influence zone should by definition not be relevant (for an assumed $\epsilon_{\mathrm{max}}$). On the other hand, zero-padding allows the same neural network to predict along system edges as well as continuous beam systems of sizes smaller than the influence zone.

\subsection{Generating an appropriate dataset}
\label{sec:method_data_generation}
As explained previously, the maximum influence zone $k_\mathrm{max}$ size depends on the design constraints and an assumed error threshold $\epsilon_{\mathrm{max}}$. These design constraints can be defined by setting minimum and maximum ranges on the known priors, UDLs $\bm{\omega}$ and spans $\mathbf{L}$, as well as the cross-section properties within vector $\mathbf{P}=[I, A_{z}, W_{\mathrm{pl}}]$:

\begin{equation}
    \label{eq:design_constraints}
    \begin{array}{r@{\ }c@{\ }l}
        \omega_\mathrm{min} & < \omega_i & < \omega_\mathrm{max} \\
        L_\mathrm{min} & < L_i & < L_\mathrm{max} \\
        I_{\mathrm{min}} & < I_i & < I_{\mathrm{max}} \\
        A_{z,\mathrm{min}} & < A_{z,i} & < A_{z,\mathrm{max}} \\
        W_{\mathrm{pl},\mathrm{min}} & < W_{\mathrm{pl},i} & < W_{\mathrm{pl},\mathrm{max}}
    \end{array}
\end{equation}

Constraints for each of these variables were chosen generously to cover the entire range of potential continuous beam systems that arise in structural design (from fixed framed multi-storey buildings to continuous bridge decks). Table \ref{table:load_span_limits} highlights the ranges chosen for the UDLs and spans, along with the interval at which these inputs were sampled at using a random uniform distribution.

\begin{table}[htb]
    \centering
    \begin{tabular}{cccc}
        \toprule
        Property & Min. & Interval & Max. \\ \midrule
        $\omega$ {\small [kN/m]} & 5 & 5 & 325 \\ 
        $L$ {\small [m]} & 0.5 & 0.5 & 20.0 \\ 
        \bottomrule
    \end{tabular}
    \caption{Ranges and intervals of know priors used for the influence zone evaluation and data generation.}
    \label{table:load_span_limits}
\end{table}

Although arbitrary cross-section property combinations could have been chosen for $I$, $A_{z}$ and $W_{\mathrm{pl}}$, using cross-section properties from an explicitly defined set ensures the predicted cross-section properties are physically realistic. Initially, the standardised UB cross-sections from BS EN 10365 \cite{CEN_2017_BS_EN_10365_2017} were considered. However, the minimum and maximum cross-section properties from this set were not sufficient for the lightest and heaviest loading conditions possible under the design constraints set by Table \ref{table:load_span_limits}. For this reason, a set of custom I-sections were generated and used exclusively for all members.

These custom I-sections were generated by averaging the geometrical ratios between the web depth $d_\mathrm{w}$, flange thickness $t_\mathrm{f}$, flange breadth $b_\mathrm{f}$ and the web thickness $t_\mathrm{w}$ that arise in BS EN 10365 \cite{CEN_2017_BS_EN_10365_2017}. Aside from ensuring that they share commonalities with the UB BS EN 10365, this process also ensured at minimum Class 2 sections \cite{CEN_2015_BS_EN_1993_1_1_2005_A1_2014} to allow the use of plastic cross-section properties. 1000 individual cross-sections were generated that ensured equal spacing across these ratios. The resulting granularity (as opposed to the 91 within BS EN 10365) meant that the utilisation ratio precision achievable during data-generation was significantly higher. The custom I-sections and associated cross-section properties are shown in Table \ref{table:XS_comparison}.

\begin{table}[htb]
    \centering
    \begin{tabular}{ccccccccc}
        \toprule
        
        \multirow[c]{2}{1.4cm}[-0.05cm]{\centering Diagram} & \multirow[c]{2}{1.4cm}[-0.05cm]{\centering Property} &  \multicolumn{3}{c}{UB BS EN10365} & & \multicolumn{3}{c}{Custom I-sections} \\ \cmidrule{3-5} \cmidrule{7-9}
        & & Min & Mean & Max & & Min & Mean & Max \\ \midrule
        \multirow[c]{10}{2.5cm}[0.05cm]{\hspace*{-0.8cm}\includegraphics[width=0.29\textwidth,height=\textheight,keepaspectratio]{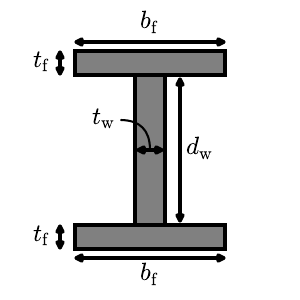}} & $t_\mathrm{w}$ {\small [mm]} & 4.0 & 13.0 & 36.1 & & 3.0 & 24.0 & 45.0 \\ 
        & $d_\mathrm{w}$ {\small [mm]} & 112 & 550 & 928 & & 30.3 & 243 & 455 \\ 
        & $t_\mathrm{f}$ {\small [mm]} & 6.8 & 21.4 & 65.0 & & 4.8 & 38.6 & 72.4 \\ 
        & $b_\mathrm{f}$ {\small [mm]} & 76.0 & 220 & 421 & & 55.5 & 444 & 833 \\ \cmidrule{2-9}
        & $d_\mathrm{w} / t_\mathrm{w}$ & 23.9 & 43.9 & 59.9 & & 43.9 & 43.9 & 43.9 \\
        & $t_\mathrm{f} / t_\mathrm{w}$ & 1.19 & 1.61 & 1.90 & & 1.61 & 1.61 & 1.61 \\
        & $b_\mathrm{f} / t_\mathrm{w}$ & 8.7 & 18.5 & 27.5 & & 18.5 & 18.5 & 18.5 \\ \cmidrule{2-9}
        & $A_\mathrm{z}$ {\small [cm$^{2}$]} & 4.47 & 87.9 & 334 & & 4.1 & 329.2 & 922  \\
        
        & $I$ {\small [cm$^{4}$]} & 473 & {$233 \times 10^3$} & {$1.25 \times 10^6$} & & 305 & {3.34 $\times 10^6$} & {15.5 $\times 10^6$}  \\
        & $W_{\mathrm{pl}}$ {\small [cm$^{3}$]} & 84.2 & 6110 & {$28.0 \times 10^3$} & & 49.4 & {$44.9 \times 10^3$} & {$167 \times 10^3$} \\ \bottomrule
    \end{tabular}
    \caption{Cross-section properties comparison between Universal Beams (UB) from BS EN 10365:2017 and custom generated I-sections. Note in particular that mean dimension ratios ($d_\mathrm{w} / t_\mathrm{w}$ etc.) are identical for both groups of cross-sections.}
    \label{table:XS_comparison}
\end{table}

Together, these efforts ensure that the dataset on which the neural network is trained on covers sufficient breadth in terms of the input and output space to generalise for a wide variety of continuous beam systems. The dataset generated based on the aforementioned design constraints, the concept of influence zones, and the technique of zero-padding were chosen with the aim to maximise the generalisability of the inverse operator for any system size $m$, UDLs $\bm{\omega}$ and spans $\mathbf{L}$. This leaves only the utilisation ratios $\mathbf{u}$ as the remaining input variable in Equation \ref{eq:ML_inverse_operator}. Instead of passing utilisation ratios as explicit inputs to the network, it was decided that the dataset will be generated so that all beams closely correspond to the target utilisation ratio $u_{\mathrm{target}}$. The network will therefore implicitly learn the $u_{\mathrm{target}}$ from the data itself.

The dataset was generated by designing continuous beam systems of size $m=2 k_\mathrm{max} + 1$ with each member having a span $L$ and UDL $\omega$ value drawn from a random uniform distribution based on the discretised ranges and intervals specified in Table \ref{table:load_span_limits}. These heterogeneous structural systems were modelled and optimised using third-party software (Rhino3D\textsuperscript{\tiny\textcopyright}, Grasshopper\textsuperscript{\tiny\textcopyright} and Karamba3D\textsuperscript{\tiny\textcopyright} \cite{PreisingerC_2013_Linking_Structure_and}) after having identified the influence zone $k_\mathrm{max}$ for the design constraints in Table \ref{table:load_span_limits} and Table \ref{table:XS_comparison}. The beams were optimised for minimum depth against ULS cross-section checks from EN 1993-1-1 6.2 \cite{CEN_2015_BS_EN_1993_1_1_2005_A1_2014} using a coupled analysis and design procedure \cite{SakaM_etal_2013_Mathematical_and_Metaheuristic} with a target utilisation ratio of $u_{\mathrm{target}}= 0.99$.

\subsection{Neural network training procedure}

\begin{figure}[htb]
    \centering
    \includegraphics[width=\textwidth,height=\textheight,keepaspectratio]{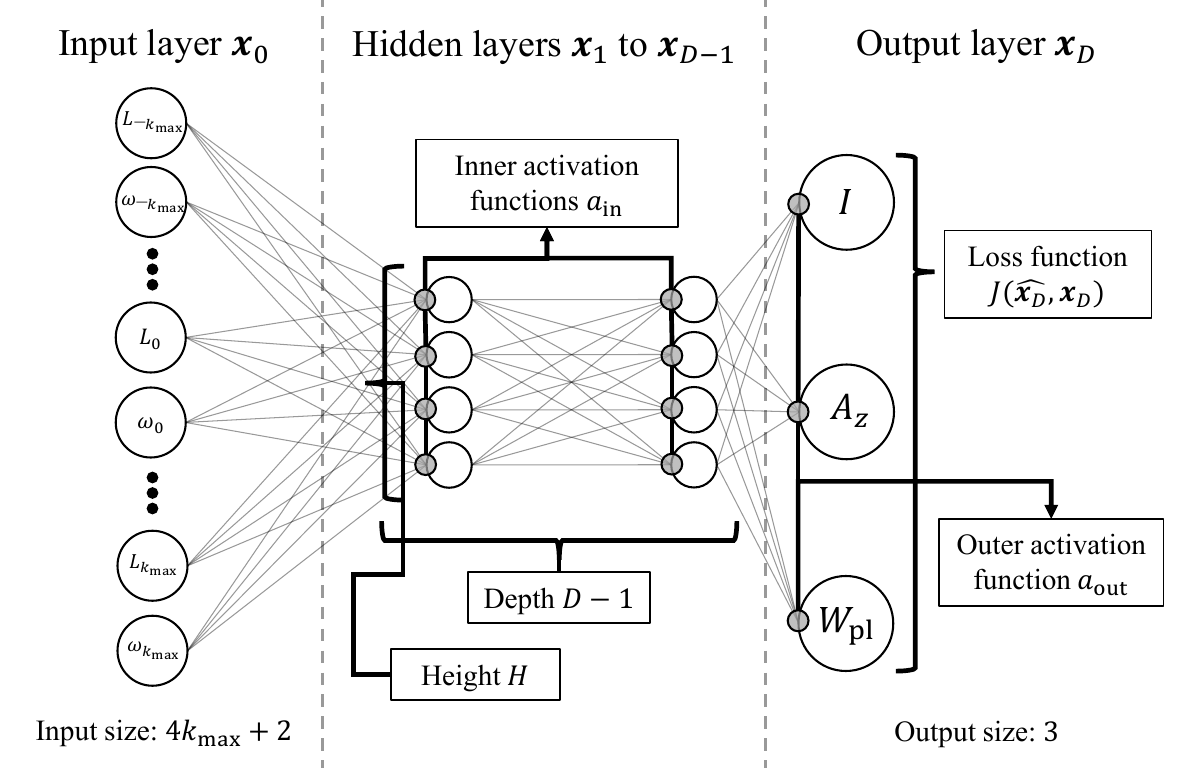}
    \caption{Generalised neural network structure with known priors from the influence zone $k_\mathrm{max}$ as the input layer $\mathbf{x}_0$ and cross-section properties of the beam as the output layer $\mathbf{x}_D$.}
    \label{fig:NN_structure}
\end{figure}

The generalised neural network structure developed in this work is shown in Figure \ref{fig:NN_structure}. Identifying an appropriate architecture in terms of height $H$, depth $D$ and activation functions $a_d$ requires experimentation. The choice of loss function $J$ to compare predicted targets $\mathbf{\widehat{x}}_D$ against true targets $\mathbf{x}_D$ also form part of the experimentation process.

\subsubsection*{Loss functions and performance metrics}
\label{sec:method_loss_functions}
In this study, four different loss functions were investigated as shown in Table \ref{table:loss_functions}. These include the Mean Absolute Error (MAE) and Mean Square Error (MSE) loss functions that are commonly used for regression models. One limitation associated with both is that their derivates (in respect to predicted targets) back-propagate the model parameters $\bm{\theta}$ with no regards what the relative size of the error is in relation to the magnitude of the output variables $I$, $A_{z}$ and $W_{\mathrm{pl}}$.

\begin{table}[htb]
    \centering
    \begin{tabular}{ccc}
        \toprule
        \multirow[c]{1}{3cm}[0.0cm]{\centering Name} & Formula & Derivative \\ \midrule
        
        \multirow[c]{1}{3cm}[0.2cm]{\centering Mean Absolute Error} &
        $ \displaystyle{J_{\text{MAE}} = \frac{1}{p} \sum_{i=1}^{p}
            \left|\widehat{x_{D,i}} - x_{D,i} \right|}$ & 
        $ \displaystyle{{J_{\text{MAE}}} ' = \frac{1}{p} \sum_{i=1}^{p}
            \begin{cases}
                \text{-}1, & \widehat{x_{D,i}} < x_{D,i} \\
                \;1, & \widehat{x_{D,i}} > x_{D,i} \\
        \end{cases} }$ \\ [.8cm]
        
        \multirow[c]{1}{3cm}[0.2cm]{\centering Mean Squared Error} &
        $\displaystyle{J_{\text{MSE}} = \frac{1}{2p} \sum_{i=1}^{p}
            (\widehat{x_{D,i}} - x_{D,i})^2}$ &
        $\displaystyle{{J_{\text{MSE}}}' = \frac{1}{p} \sum_{i=1}^{p}
            \widehat{x_{D,i}} - x_{D,i}}$ \\ [.8cm]
        
        \multirow[c]{1}{3cm}[0.2cm]{\centering Mean Absolute Percentage Error} &
        $ \displaystyle{J_{\text{MAPE}} = \frac{1}{p} \sum_{i=1}^{p} \left| \frac{\widehat{x_{D,i}} - x_{D,i}}{x_{D,i} + \epsilon} \right|}$ &
        $ \displaystyle{{J_{\text{MAPE}}}' = \frac{1}{p} \sum_{i=1}^{p}
            \begin{cases}
                \frac{\text{-}1}{|x_{D,i} + \epsilon|}, & \widehat{x_{D,i}} < x_{D,i} \\
                \frac{1}{|x_{D,i} + \epsilon|}, & \widehat{x_{D,i}} > x_{D,i} \\
        \end{cases} }$ \\ [.8cm]
        
        \multirow[c]{1}{3cm}[0.2cm]{\centering Mean Squared Percentage Error} &
        $\displaystyle{J_{\text{MSPE}} = \frac{1}{2p} \sum_{i=1}^{p}
            \frac{(\widehat{x_{D,i}} - x_{D,i})^2}{|x_{D,i} + \epsilon|}}$ &
        $\displaystyle{{J_{\text{MSPE}}}' = \frac{1}{p} \sum_{i=1}^{p}
            \frac{\widehat{x_{D,i}} - x_{D,i}}{|x_{D,i} + \epsilon|}}$ \\
        \bottomrule
    \end{tabular}
    \caption{Loss functions to be tested with $p$ predicted targets $\widehat{x}_D$, $p$ true targets $x_D$ and small $\epsilon$ to avoid division by zero errors.}
    \label{table:loss_functions}
\end{table}

This is problematic given the orders of magnitude difference between the largest and smallest section properties of the custom I-sections as shown in Table \ref{table:XS_comparison}. An error of $100~\mathrm{cm}^4$ for $I$ would cause the same back propagation adjustment using MAE or MSE regardless if the true second moment of area value target is $305~\mathrm{cm}^4$ or $305 \times 10^5~\mathrm{cm}^4$. As a consequence, both MAE and MSE would prioritise minimising the absolute error, which mathematically favours target values of large magnitudes at the expense of smaller ones.

To address the above mentioned issue, percentage-based versions of both MAE and MSE were tested, defined in Table \ref{table:loss_functions} as the Mean Absolute Percentage Error (MAPE) and the Mean Squared Percentage Error (MSPE). Whilst MAPE is commonly used, MSPE is not tested in practice. Both MAPE and MSPE ensure that during back-propagation, the optimiser updates model parameters in proportion to the relative deviation between predicted $\mathbf{\widehat{x}}_D$ and true outputs $\mathbf{x}_D$, which should be a better performance criterion to address the orders of magnitude difference in the output space that arise in these particular continuous beam systems.

Regardless of the choice of loss function, MAPE will be used as a comparison metric between different networks. However to study the dispersion of prediction errors, an accuracy metric $M$ will also be evaluated with minimum, 0.5\%, 2.5\%, 50\% (median), 97.5\%, 99.5\% and maximum percentile values. This will allow the evaluation of the 95\% and 99\% confidence intervals (CI) and help identify the range of over and under prediction of outputs, which is important in the context of safe structural design:

\begin{equation}
    \label{eq:performance_metrics}
    M(\widehat{\mathbf{x}_D},\mathbf{x}_D) = \frac{1}{p} \sum_{i=1}^{p} \frac{\widehat{x_{D,i}}} {x_{D,i}} = \frac{1}{3} \left( \frac{\widehat{I}}{I} + \frac{\widehat{A_{z}}}{A_{z}} +\frac{\widehat{W_{\mathrm{pl}}}}{W_{\mathrm{pl}}} \right).
\end{equation}

\subsubsection*{Activation functions}
The activation functions tested in this work are listed in Table \ref{table:activation_functions}, and includes the commonly used rectified linear unit (ReLU) function amongst others \cite{MurphyK_2022_Probabilistic_machine_learning_}. A distinction is drawn between the inner activation functions $a_{\mathrm{in}}$ within the hidden layers, and the outer activation function $a_{\mathrm{out}}$, that evaluate the target values $\mathbf{x}_D$. All inputs and outputs were scaled between 0 and 1 by dividing the values by the maximum magnitude of the features and targets within the training and validation set, respectively. Therefore, it is important to choose only output activation functions compatible with the scaled values of the targets as reflected by the range of $a_{\mathrm{out}}$ functions listed in Table \ref{table:activation_functions}.


\begin{table}[htb]
    \centering
    \begin{tabular}{cccc}
        \toprule
        Name & Formula & $a_{\mathrm{in}}$ & $a_{\mathrm{out}}$ \\ \midrule
        
        Rectified linear unit (ReLU) &
        $\displaystyle{a_{\mathrm{ReLU}} = \mathrm{max}(\mathbf{wx + b}, 0)}$ & \checkmark & \checkmark \\
        
        Sigmoid &
        $\displaystyle{a_{\mathrm{sigm}} = 1 / (1+e^\mathbf{wx + b})}$ & \checkmark & \checkmark \\
        
        Hyperbolic tangent &
        $\displaystyle{a_{\mathrm{tanh}} = \mathrm{tanh}(\mathbf{wx + b})}$ & \checkmark & \\ 
        
        Exponential &
        $\displaystyle{a_{\mathrm{exp}}=e^\mathbf{wx + b}}$ & & \checkmark \\
        
        \bottomrule
    \end{tabular}
    \caption{Table of inner $a_{\mathrm{in}}$ and outer $a_{\mathrm{out}}$ activation functions to be tested with weight vector $\mathbf{w}$, bias vector $\mathbf{b}$ and layer vector $\mathbf{x}$.}
    \label{table:activation_functions}
\end{table}

\subsubsection*{Height and depth analysis}
\label{sec:method_breadth_and_depth}
The appropriate size of a neural network in terms of height $H$ and depth $D$ was found by finding a suitable trade-off between under and over fitting the model parameter space. The design complexity of continuous beam systems will likely be reflected in deeper and wider neural networks than those considered in previous literature \cite{TseranidisS_etal_2016_Data_driven_approximation_algorithms} due to the large number of load arrangements that may be critical, the numerous design criteria that govern the design, and the variety of viable cross-sections. For this reason, a wide range of heights and depths were tested. The size of the networks were denoted by a simple syntax based on the architecture of the hidden layers. For example, ``50-50-50'' refers to a neural network with three hidden layers with 50 nodes each.

\subsubsection*{Other neural network parameters and hyperparameters}
\label{sec:method_other_parameters}

Given the large dataset size and computational resources required for training, a simple hold-out strategy was deemed appropriate as opposed to other validation strategies \cite{KimJ_2009_Estimating_classification_error}, and hence the final dataset was randomised and split into training, validation and testing sets using a 70\%, 15\%, 15\% split, respectively. The testing set was only used once after an appropriate neural network architecture was found experimentally. A robustness check with various initialiser seeds was carried out on the final architecture. Other neural network training aspects, such as optimisers, types of initialisers, learning rates and batch-sizes were chosen empirically based on MAPE performance and qualitative comparison of learning behaviour. The options/ranges for each of these are summarised in Table \ref{table:other_NN_parameters}. All stochastic elements were controlled through explicit initialiser seeds.

\begin{table}[htb]
    \centering
    \begin{tabular}{ccc}
        \toprule
        Neural network training aspect & Options/range considered & Selected \\ \midrule
        
        Optimiser &
        SGD, RMSprop, Adam, Nadam &
        Nadam \\
        
        Learning rate &
        $\alpha = [0.0001, 1.0000]$ &
        $\alpha$ = 0.0005 \\
        
        Initialiser &
        Gaussian with $\mu = 0$, $\sigma = [0.00, 1.00]$&
        $\sigma = 0.05$\\ 
        
        Batch-size &
        $[128,  8192]$&
        1024 \\
        \bottomrule
    \end{tabular}
    \caption{Options and/or ranges of neural network learning parameters and hyperparameters tested, along with selected parameters for all training runs presented in results.}
    \label{table:other_NN_parameters}
\end{table}

\subsection{Summary of methodology}

The following procedure was adopted to develop the machine learned inverse operator:

\begin{enumerate}
    \item Evaluate the maximum influence zone $k_\mathrm{max}$ for the continuous beam system using the procedure from \cite{GalletA_etal_2023_Influence_zones_for} based on the design constraints specified in Tables \ref{table:load_span_limits} and \ref{table:XS_comparison}.
    \item Design continuous beam systems of size $m = 2 k_\mathrm{max} + 1$ using a coupled analysis and design approach \cite{SakaM_etal_2013_Mathematical_and_Metaheuristic} with a target utilisation ratio $u_{\mathrm{target}}= 0.99$. Each beam within the continuous beam system will correspond with one data point, with zero-padding for edge or near-edge beams as shown in Figure \ref{fig:zero_paddding_inputs}. Finally, split and normalise the data into a training, validation and testing set as explained in Section \ref{sec:method_other_parameters}.
    \item Develop the neural network model using the following steps:
    \begin{enumerate}
        \item Assume a standard 50-50 architecture and test out the various combinations of loss and activation functions as identified Table \ref{table:loss_functions} and Table \ref{table:activation_functions}, respectively, based on 100k training data points.
        \item Test various height $H$ and depth $D$ variations as explained in Section \ref{sec:method_breadth_and_depth} based on 100k training data points.
        \item For the best architecture (height, depth and activation function), test the performance against different training set sizes.
    \end{enumerate}
    \item Evaluate the performance of the final neural network against the testing dataset and conduct a robustness test using various initialiser seeds for the weights and biases. 
\end{enumerate}

%% file: 04_Results.tex
\section{Model development and results}
\label{sec:results}

\subsection{Influence zone size estimation}
\label{sec:results_influence_zone}

\begin{figure}[!htb]
    \centering
    \includegraphics[width=0.63\textwidth,height=\textheight,keepaspectratio]{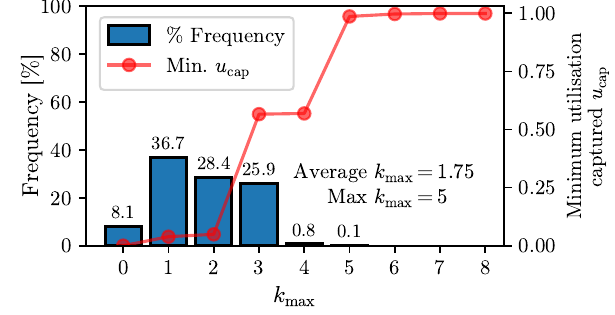}
    \caption{Influence zone results for a $m=17$ system with $\epsilon_{\mathrm{max}} = 0.02$ based on the design constraints established by Tables \ref{table:load_span_limits} and \ref{table:XS_comparison} using the methodology from \cite{GalletA_etal_2023_Influence_zones_for}.}
    \label{fig:results_influence_zone}
\end{figure}

The maximum influence zone $k_{\mathrm{max}}$ of continuous beam systems subject to design constraints specified by Tables \ref{table:load_span_limits} and \ref{table:XS_comparison} was established.
Using the procedure from \cite{GalletA_etal_2023_Influence_zones_for}, 25 random UDL and span distributions were generated for a $m=17$ sized system and designed against ULS checks from EN 1993-1-1 \cite{CEN_2015_BS_EN_1993_1_1_2005_A1_2014} using the custom I-sections specified in Table \ref{table:XS_comparison} with a target utilisation ratio $u_{\mathrm{target}}= 0.99$. This led to the creation of 10,625 continuous beams ($25 \times 25 \times 17$). Each beam's influence zone value was evaluated using an error threshold of $\epsilon_{\mathrm{max}}=0.02$. This threshold was selected based on the expected MAPE performance achievable with the multi-layered neural network, with the results shown in Figure \ref{fig:results_influence_zone}. The results indicate that the average and maximum influence zone size is $k_{\mathrm{max}}=1.75$ and $k_{\mathrm{max}}=5$, respectively. This suggests the system size required for the dataset generation is $m=2 k_\mathrm{max} + 1 = 11$, and the required input layer size is $4 k_\mathrm{max} + 2 = 22$. This influence zone evaluation took 5 hours of computation time.

\subsection{Data generation, visualisation and pre-processing}
\label{sec:results_data_generation_prepocessing}
Drawing from uniform distributions for spans and UDL values identified in Table \ref{table:load_span_limits}, two datasets were created. The first consists out of 266 unique UDL $\bm{\omega}$ and span $\mathbf{L}$ permutations of $m=11$ sized continuous beam systems, and the second out of 251 unique permutations. A coupled analysis and design optimisation approach \cite{SakaM_etal_2013_Mathematical_and_Metaheuristic} with a target utilisation ratio $u_{\mathrm{target}}= 0.99$ based on ULS cross-section checks \cite{CEN_2015_BS_EN_1993_1_1_2005_A1_2014} and all critical load arrangements \cite{GalletA_etal_2023_Influence_zones_for} was implemented to find the appropriate custom I-section from Table \ref{table:XS_comparison} for each beam within the system. This process resulted in 1,471,327 individual data-points ($11 \times (266^2 + 251^2)$) that took 3.5 days to generate.

\begin{figure}[htb]
    \centering
    \includegraphics[width=\textwidth,height=\textheight,keepaspectratio]{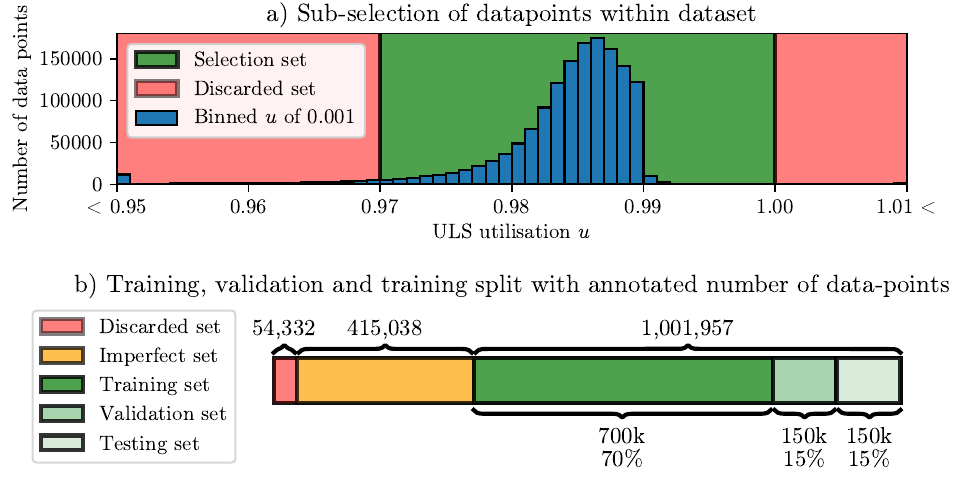}
    \caption{Sub-selection of data points from the initial 1,471,327 dataset based on a ULS utilisation ratio $u$ range of 0.97-1.00. The 1000k CBeamXP dataset was drawn from the green sets.}
    \label{fig:u_selection}
\end{figure}

The distribution of utilisation ratios achieved for the specified target utilisation ratio $u_{\mathrm{target}}= 0.99$ are shown in Figure \ref{fig:u_selection}a). Since the design space is limited to discretized cross-section properties the utilisation target ratio $u_{\mathrm{target}}= 0.99$ was rarely met exactly. Therefore, a sub-selection of this dataset took place, discarding all of the beams that fell outside of utilisation ratio range $0.97 \le u < 1.00$.

Although only 54,322 data-points belonged to the discarded set, the dataset was further stripped of all data points which had beam members within their system that belonged to the discarded set, even if those beams themselves fell within the selected utilisation ratio range. These data points are defined as the ``Imperfect set'' in Figure \ref{fig:u_selection}a). This process removed another 415,038 data-points. This left 1,001,957 data points (1,471,327-415,038-54,322), each representing a beam within a $m=11$ member system and the surrounding design information from the influence zone of valid structural designs under ULS conditions. This set was randomised, further stripped of another 1957 data points, to yield a dataset size of exactly 1 million (1000k).

\begin{figure}[htb]
    \centering
    \includegraphics[width=\textwidth,height=\textheight,keepaspectratio]{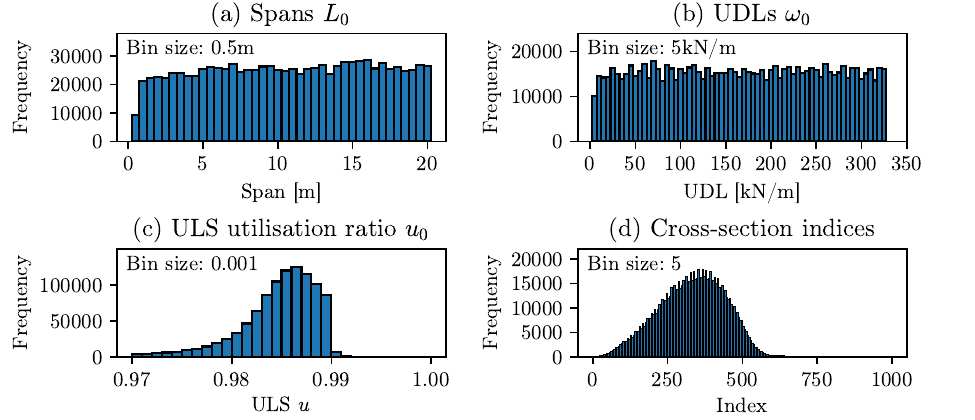}
    \caption{Frequency distributions for various descriptor variables of the CBeamXP dataset. Spans and UDL values are uniformly distributed, whilst the selected cross-section indices of the optimised beam systems follow a normal distribution.}
    \label{fig:final_dataset}
\end{figure}

This 1000k dataset was named CBeamXP: Continuous Beam Cross-section Predictors \cite{GalletA_etal_2023_CBeamXP__Continuous_Beam} and represents ULS compliant beam systems of system size $m=11$ with utilisation ratios between $0.97 \le u < 1.00$. The CBeamXP dataset was split into a training, validation and testing set using a 70\%, 15\%, 15\% split as shown in Figure \ref{fig:u_selection}b). Histograms of the spans, UDLs, utilisation ratios and cross-section indices (corresponding to one of the 1000 custom I-sections in ascending stiffness order) are shown in Figure \ref{fig:final_dataset}. For pre-processing, all inputs and outputs were divided by the maximum value within the training and validation set.

\subsection{Network architecture development results}

\subsubsection{Loss and activation function variations}
\label{sec:results_loss_and_activation}
The neural network development began by evaluating the MAPE performance of 50-50 architectures for different loss functions $J$, inner $a_\mathrm{in}$ and outer $a_\mathrm{out}$ activation functions. These networks were trained for 1000 epochs using 100k datapoints from the training set, yet validated against the entire 150k validation set. The total training time was 5 hours with results shown in Table \ref{table:results_50_50_loss_and_activation_comparison}.

\newcommand{\BC}[1]{{\textbf{#1}}} 
\begin{table}[!htb]
    \centering
    \small
    \begin{tabular}{c ccc c ccc}
        \toprule
        & \multicolumn{3}{c}{$J_\mathrm{MAE}$} & ~ & \multicolumn{3}{c}{$J_\mathrm{MSE}$} \\ 
        & $a_\mathrm{out,ReLU}$ & $a_\mathrm{out,sigm}$ & $a_\mathrm{out,exp}$ & ~ & $a_\mathrm{out,ReLU}$ & $a_\mathrm{out,sigm}$ & $a_\mathrm{out,exp}$ \\ \cmidrule{2-4} \cmidrule{6-8}
        $a_\mathrm{in,ReLU}$ & 0.105 & 0.164 & 0.143 & ~ & 0.116 & 0.168 & 0.146 \\
        $a_\mathrm{in,sigm}$ & 0.558 & 0.158 & 0.168 & ~ & 0.568 & 0.250 & 0.243 \\
        $a_\mathrm{in,tanh}$ & 0.205 & 0.143 & 0.133 & ~ & 0.274 & 0.160 & 0.147 \\ \\
        
        & \multicolumn{3}{c}{$J_\mathrm{MAPE}$} & & \multicolumn{3}{c}{$J_\mathrm{MSPE}$} \\ 
        & $a_\mathrm{out,ReLU}$ & $a_\mathrm{out,sigm}$ & $a_\mathrm{out,exp}$ & & $a_\mathrm{out,ReLU}$ & $a_\mathrm{out,sigm}$ & $a_\mathrm{out,exp}$ \\ \cmidrule{2-4} \cmidrule{6-8}
        $a_\mathrm{in,ReLU}$ & \BC{0.072} & \BC{0.090} & \BC{0.091} & ~ & \BC{0.089} & \BC{0.093} & \BC{0.089} \\
        $a_\mathrm{in,sigm}$ & 0.707 & 0.129 & 0.124 & ~ & 0.707 & 0.147 & 0.137 \\
        $a_\mathrm{in,tanh}$ & 0.205 & \BC{0.083} & \BC{0.083} & ~ & 0.235 & \BC{0.082} & \BC{0.087} \\
        
        \bottomrule
    \end{tabular}
    \caption{Validation MAPE metrics at epoch 1000 for different combinations of loss $J$, inner $a_\mathrm{in}$ and outer $a_\mathrm{out}$ activation functions for a 50-50 architecture using 100k training and 150k validation data points. MAPE values of less than 0.100 (10\%) are in bold.}
    \label{table:results_50_50_loss_and_activation_comparison}
\end{table}

The results clearly indicate that the percentage-based loss functions $J_\mathrm{MAPE}$ and $J_\mathrm{MSPE}$ typically outperform their non-percentage-based counter-parts. Performance between either $J_\mathrm{MAPE}$ and $J_\mathrm{MSPE}$ was relatively similar, with all MAPE values of less than 10\% (0.100) bolded. $J_\mathrm{MAPE}$ was chosen as the loss function for this investigation due to it being more commonly used. The five best inner and outer activation function combinations in Table \ref{table:results_50_50_loss_and_activation_comparison} under $J_\mathrm{MAPE}$ (bolded values) were subsequently qualitatively analysed.

This qualitative analysis highlighted that using ReLU as an outer activation function allows the prediction of null cross-section properties. This is an invalid prediction since the model operates on the basis that a beam with some minimum cross-section properties must exist in the context of this structural system. A similar limitation applied for the sigmoid activation function which asymptotically approaches the value of one at positive infinity. This limits the network's ability of predicting cross-sections larger than those found within the training and validation dataset. For these reasons, the $a_\mathrm{out,exp}$ function was selected for this particular network architecture since $a_\mathrm{out,exp}$ does not result in zero-valued cross-section properties and also does not impose an upper limit on the outputs. From the remaining viable networks, the $a_\mathrm{in,ReLU}$ and $a_\mathrm{out,exp}$ architecture converged the quickest and was therefore chosen for further development in this study.

\subsubsection{Height and depth variations}
\label{sec:results_height_and_depth}
\begin{figure}[!htb]
    \centering
    \includegraphics[width=\textwidth,height=\textheight,keepaspectratio]{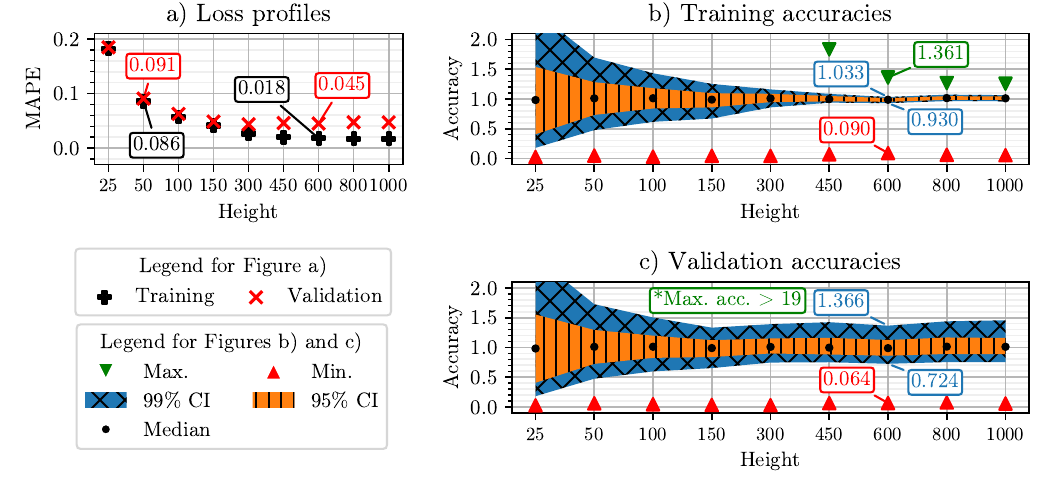}
    \caption{Loss and accuracy profiles for $a_\mathrm{in,ReLU}$ and $a_\mathrm{out,exp}$ networks at epoch 1000 with $J_\mathrm{MAPE}$ with two hidden layers of equal height. Training set size of 100k and validation set size of 150k.}
    \label{fig:results_height_analysis}
\end{figure}

The architecture of the hidden layers (height $H$ and depth $D$) needs to be sufficiently expressive to reflect the design complexity of continuous beam systems, and need to avoid under- and over-fitting the model. Therefore, comparison between training and validation performance is needed. The 100k training and 150k validation sets from section \ref{sec:results_loss_and_activation} were re-used for this purpose. Figure \ref{fig:results_height_analysis} compares the performance of various networks containing two hidden layers of varying heights at epoch 1000. The combined training time of these networks was 22 hours. The 600-600 network was identified as the point at which the performance transitioned from under-fitting to slight over-fitting. Figures \ref{fig:results_height_analysis}b) and c) further indicate the accuracy profiles for both training and validation, respectively. Note that the maximum validation accuracy values greatly exceed the value of at least 19 (1900\%) for all networks, regardless of height, meaning the network predicted cross-sectional properties 19 times larger than the target value.

\begin{figure}[!htb]
    \centering
    \includegraphics[width=\textwidth,height=\textheight,keepaspectratio]{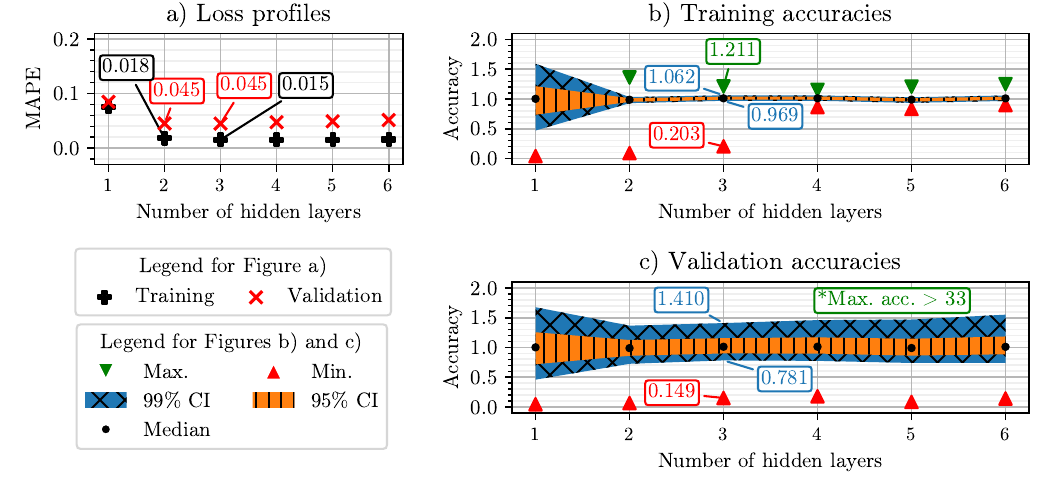}
    \caption{Loss and accuracy profiles for $a_\mathrm{in,ReLU}$ and $a_\mathrm{out,exp}$ networks at epoch 1000 with $J_\mathrm{MAPE}$ with hidden layers of height $H=600$. Training set size of 100k and validation set size of 150k.}
    \label{fig:results_depth_analysis}
\end{figure}

The ``optimal'' number of hidden layers for height 600 was investigated, with results shown in Figure \ref{fig:results_depth_analysis}. This resulted in a combined training time of 12 hours. Networks with more than three hidden layers showed no major improvements in either training or validation performance except in minimum training accuracy. However, this was not associated with an improvement in minimum validation accuracy as shown in Figure \ref{fig:results_depth_analysis}c). For these reasons, a depth of three hidden layers was deemed appropriate.

\subsubsection{Dataset size variations}
\label{sec:results_dataset_size}
\begin{figure}[!htb]
    \centering
    \includegraphics[width=\textwidth,height=\textheight,keepaspectratio]{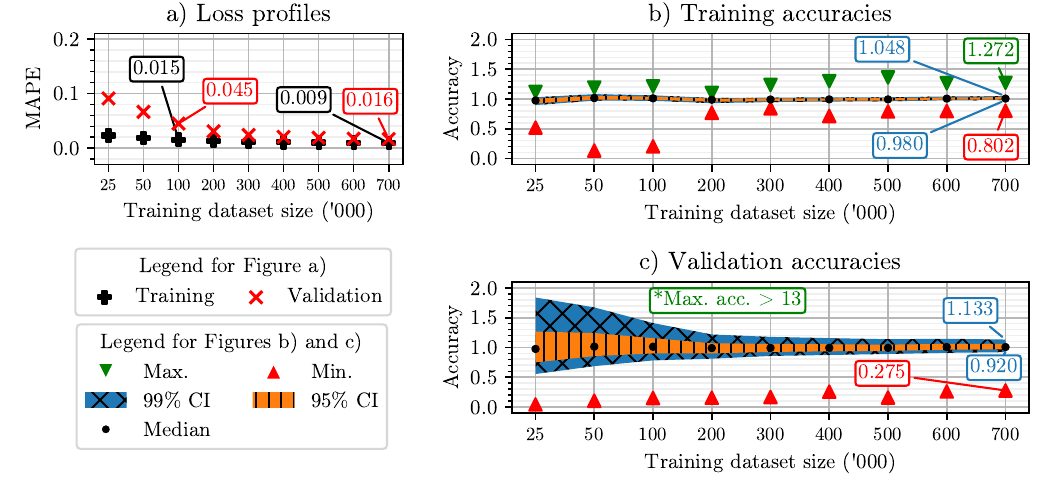}
    \caption{Loss and accuracy profiles for 600-600-600 $a_\mathrm{in,ReLU}$ and $a_\mathrm{out,exp}$ network at epoch 1000 with $J_\mathrm{MAPE}$ for various training dataset sizes, with a validation set size of 150k.}
    \label{fig:results_dataset_size_analysis}
\end{figure}

Figure \ref{fig:results_dataset_size_analysis} shows the change in performance as a function of the training dataset size, from 25k to 700k data points, with the same 150k validation dataset as in the previous sections. The combined training time was 1.5 days. Except for slight variations in the minimum and maximum accuracy values, the performance of the neural network naturally improved with a larger training set.

\subsection{Model performance: testing and robustness}
\label{sec:results_final_model_and_robustness}
\begin{figure}[!htb]
    \centering
    \includegraphics[width=1.0\textwidth,height=\textheight,keepaspectratio]{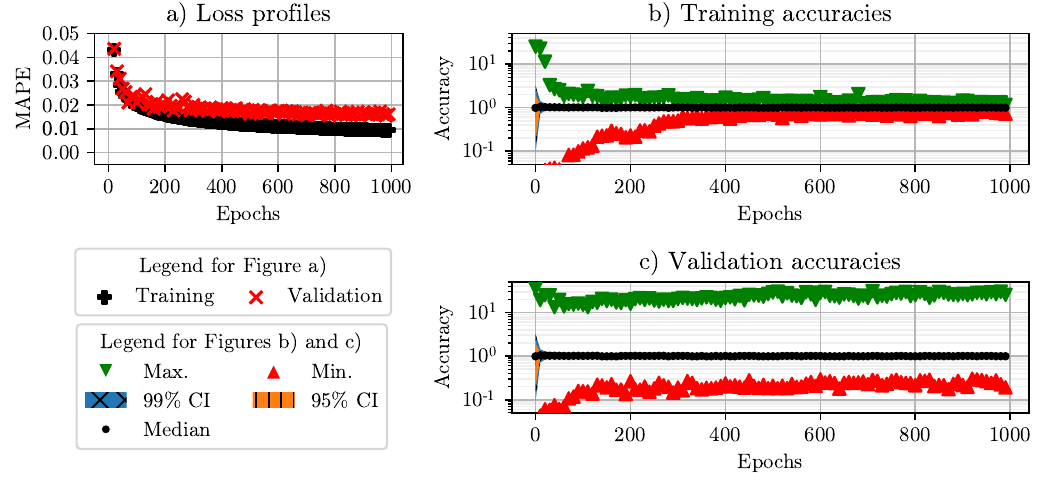}
    \caption{Loss and accuracy profiles for 600-600-600 $a_\mathrm{in,ReLU}$ and $a_\mathrm{out,exp}$ network at epoch 1000 with $J_\mathrm{MAPE}$, 700k training and 150k validation sets. Note use of logarithmic y-axis to show the full range of maximum validation accuracies.}
    \label{fig:results_final_model}
\end{figure}

The final neural network model consists of a 600-600-600 architecture with $a_\mathrm{in,ReLU}$ and $a_\mathrm{out,exp}$ activation functions trained using the  $J_\mathrm{MAPE}$ loss function based on a training and validation dataset size of 700k and 150k data points, respectively, with learning graphs shown in Figure \ref{fig:results_final_model}. The neural network at epoch 1000 was also evaluated against the testing set created in Figure \ref{fig:u_selection}, and checked for robustness by re-training the same network a further 9 times using different kernel initialiser seeds, which took 4.5 days. The general model performance results and standard deviations $\sigma_{\text{initialiser}}$ due to these different initialiser seeds is summarised in Table \ref{table:results_final_model}.

\begin{table}[!htb]
    \centering
    \footnotesize
    \begin{tabular}{cccccccccc}
        \toprule
        \multirow[b]{2}{1.5cm}[-0.1cm]{\centering Dataset} &
        \multirow[b]{2}{1.1cm}[-0.1cm]{\centering Data Type} &
        \multirow[b]{2}{1.1cm}[0.1cm]{\centering MAPE} &
        \multicolumn{7}{c}{Accuracy Percentiles} \\ \cmidrule{4-10}
        
        & & & Min & 0.5\% & 2.5\% & Median & 97.5\% & 99.5\% & Max \\ \midrule
        
        \multirow[c]{3}{1.5cm}[0.2cm]{\centering Training} & Results & 0.009 & 0.802 & 0.980 & 0.990 & 1.007 & 1.030 & 1.048 & 1.272 \\
        & $\sigma_{\text{initialiser}}$ &  0.000 & 0.161 & 0.013 & 0.010 & 0.007 & 0.010 & 0.014 & 0.138 \\ [0.3cm]
        
        \multirow[c]{3}{1.5cm}[0.2cm]{\centering Validation} & Results & 0.016 & 0.275 & 0.920 & 0.968 & 1.006 & 1.058 & 1.133 & 26.811 \\
        & $\sigma_{\text{initialiser}}$ &  0.001 & 0.058 & 0.014 & 0.010 & 0.007 & 0.012 & 0.018 & 4.977  \\ [0.3cm]
        
        \multirow[c]{3}{1.5cm}[0.2cm]{\centering Testing} & Results & 0.016 & 0.313 & 0.917 & 0.967 & 1.006 & 1.058 & 1.138 & 12.282 \\
        & $\sigma_{\text{initialiser}}$ & 0.001 & 0.070 & 0.012 & 0.010 & 0.007 & 0.012 & 0.018 & 2.189 \\ \bottomrule
    \end{tabular}
    \caption{Loss and accuracy profiles for 600-600-600 $a_\mathrm{in,ReLU}$ and $a_\mathrm{out,exp}$ network at epoch 1000 with $J_\mathrm{MAPE}$, 700k training, 150k validation sets and 150k testing set.}
    \label{table:results_final_model}
\end{table}

The similar performance between the testing and validation set in Table \ref{table:results_final_model} strongly suggests that the model is likely to generalise well to new data-points. The impact of changing initialiser seed is minimal except for the minimum and maximum accuracy values.

%% file: 05_Discussion.tex
\section{Discussion}
\label{sec:discussion}

\subsection{Model generalisability}

\begin{figure}[htb]
    \centering
    \includegraphics[width=1.0\textwidth,height=\textheight,keepaspectratio]{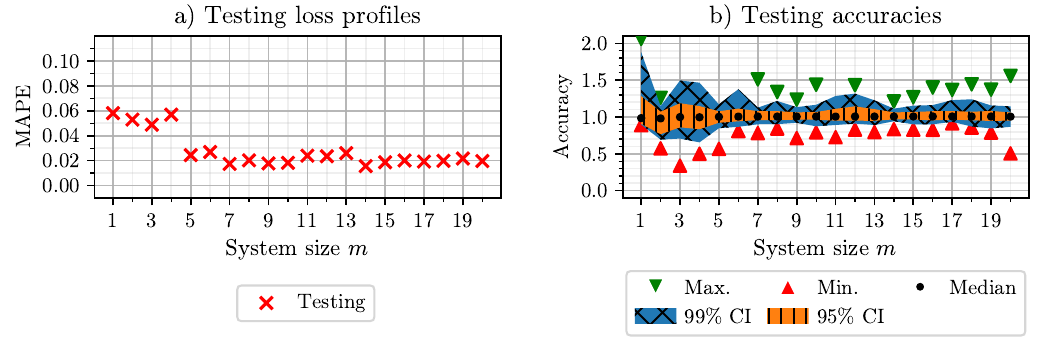}
    \caption{Generalisability performance of the final 600-600-600 neural network in terms of MAPE and accuracy for unseen structures of varying system size $m$.}
    \label{fig:discussion_model_generalisability}
\end{figure}

One of the fundamental objectives of this work was to develop a machine learned structural design model capable of generalising beyond the system size $m=11$ it was trained on. To achieve this, the influence zone concept was leveraged with zero-padding to theoretically allow the neural network to make localised predictions for continuous beams of arbitrary system size $m$. To test this, over 1000 additional testing data-points were generated using the same methodology as described in Section \ref{sec:method_data_generation} and sub-selection process as shown in Section \ref{sec:results_data_generation_prepocessing} for each system size $1 \le m \le 20$ (including $m=11$). The MAPE and accuracy performance are shown in Figure \ref{fig:discussion_model_generalisability}.

Figure \ref{fig:discussion_model_generalisability} indicates that the machine learned inverse operator demonstrates strong generalisation capability for continuous beam system sizes $m \ge 5$ with $\mathrm{MAPE} \approx 2\%$. System sizes $m<5$ saw a slightly deteriorating MAPE values of $5\% - 6\%$. These are encouraging results given that the neural network was never trained on system sizes less or greater than $m=11$. The greatest variations in performance were typically in the maximum and minimum accuracy values; in fact the model performed often better in terms of maximum performance for system size other than $m=11$.

These results provide merit to the novel implementation of the influence zone concept \cite{GalletA_etal_2023_Influence_zones_for} as a mechanics driven feature selector and using zero-padding to build machine learned inverse operators capable of generalising to differently sized continuous structural systems. This opens the possibility of investigating the applicability of this methodology for two or three dimensional frames. Furthermore, these results also provide a solution to the limitation of fixed-dimensional input vectors of multi-layer neural networks \cite{WhalenE_etal_2022_Toward_Reusable_Surrogate}.

In recent years, other researchers have investigated the development of generalisable machine learning models; most of these efforts have focused on machine learned forward operators \cite{WhalenE_etal_2022_Toward_Reusable_Surrogate, BlekerL_etal_2023_Structural_Form_Finding_Enhanced, DennisA_etal_2023_The_Direction_encoded_Neural}. Within the realm of structural design inverse operators, researchers have noted that the question of generalisability remains typically under-investigated \cite{BehzadiM_etal_2021_Real_Time_Topology_Optimization}. Whilst previous works studied the ability of neural networks to generalise under different boundary conditions \cite{XiangC_etal_2022_Accelerated_topology_optimization, NieZ_etal_2021_TopologyGAN__Topology_Optimization}, this work distinguishes itself on generalising across differently sized systems. Combining the underlying techniques behind these studies may allow one to train a generalisable model of arbitrary size and arbitrary boundary conditions.

\subsection{Performance variability}

This investigation also differentiated itself from previous works by measuring the variability of predictions in terms of accuracies. Notably, this allowed one to identify the range of over- and under-predictions, which are not captured by average loss function metrics such as MAE or MAPE. Despite gradual improvement within the 95\% and 99\% confidence intervals, the final performance graph in Figure \ref{fig:results_final_model} indicates that the confidence intervals of the validation set lag those of the training set. The same can also be said for the testing set, especially for maximum and minimum accuracies as shown in Table \ref{table:results_final_model}.

\begin{figure}[htb]
    \centering
    \includegraphics[width=1.0\textwidth,height=\textheight,keepaspectratio]{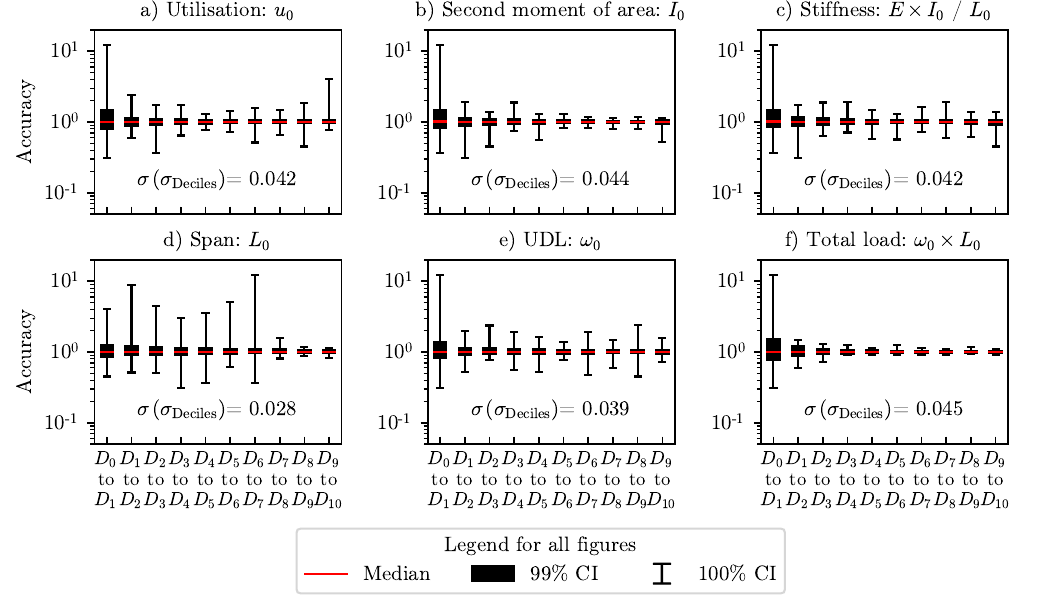}
    \caption{Custom box plots of accuracies vs. variables in binned deciles. Total load f) correlates with the greatest dispersion $\sigma (\sigma_\mathrm{Deciles}) = 0.045$, which can also be identified visually.}
    \label{fig:results_boxplots}
\end{figure}

To identify potential causes of this divergence in performance between the training and testing set, custom box plots of testing accuracies were generated for a number of variables that describe the dataset, ordered based on ascending deciles ($D_0$ to $D_{10}$). By evaluating the standard deviation of each decile's accuracy values, and taking the standard deviation of those standard deviations $\sigma (\sigma_\mathrm{Deciles})$, one can quantify numerically which variable causes the greatest dispersion of the accuracy values. These results are shown in Figure \ref{fig:results_boxplots}.

\begin{figure}[htb]
    \centering
    \includegraphics[width=0.98\textwidth,height=\textheight,keepaspectratio]{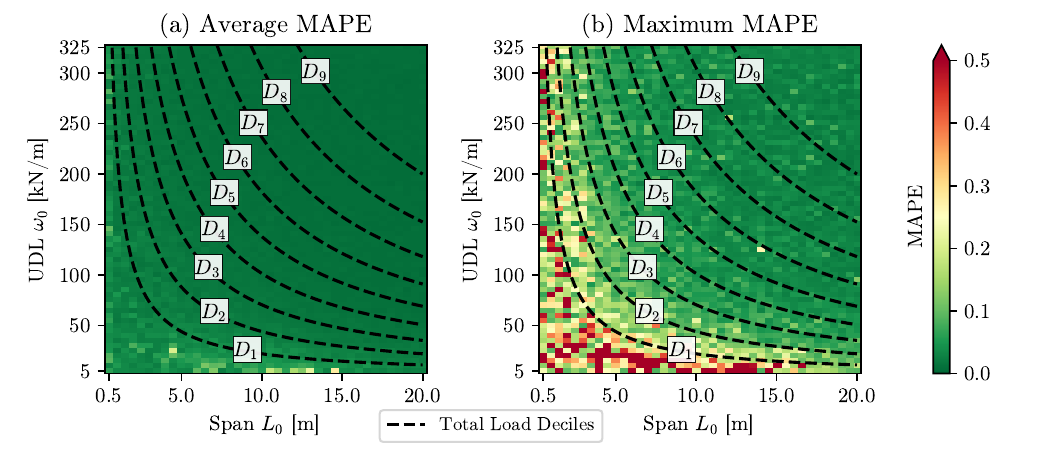}
    \caption{Total load heatmaps which evaluates the a) average MAPE and b) maximum MAPE values for all UDL $\omega_0$ and span $L_0$ combinations. Notice how maximum MAPE errors occur at low total load combinations ($\omega_0 \times L_0$) within the first decile up to $D_1$.}
    \label{fig:heatmap_total_load_deciles}
\end{figure}

By studying Figure \ref{fig:results_boxplots} in detail, it was identified that the total load variable $\omega_0 \times L_0$ caused the greatest $\sigma (\sigma_\mathrm{Deciles})$ dispersion as seen in Figure \ref{fig:results_boxplots}f). Figure \ref{fig:results_boxplots}f) also showed the most identifiable demarcation between low and high accuracy results. The prediction variability of cross-section properties of a beam is the worst when the combined product of both the UDL load $\omega_0$ and span $L_0$ fell in the lowest Decile $(< D_1)$. This pattern can also be identified by studying heat-maps of the average and maximum MAPE performance which occurred at each $\omega_0$ and span $L_0$ combination within the dataset as shown in Figure \ref{fig:heatmap_total_load_deciles}.

Using structural engineering intuition, one infers that the design of short and lightly loaded spans is more likely to be influenced by the UDLs of the surrounding members within a continuous system. Whilst the influence zone concept ensures the pertinent design information is contained within the inputs, providing that information solely in the form of an input vector may not be sufficient to make accurate predictions under all circumstances. The fact that there were also wide prediction variabilities for the smallest deciles for both the second-moment of area and stiffness values as shown in Figures \ref{fig:results_boxplots}b) and c) suggests that exposing the machine learning model to additional physics knowledge (other than influence zones) of the structural system may lead to further improvements.

In one of the early studies, Berke et al. \cite{BerkeL_etal_1993_Optimum_design_of} noted that despite achieving relatively low prediction errors on average, neural network predictions can occasionally vary significantly. In more recent works, the presence of large error predictions (over $>$40\%) were noted and manually removed from the final reported average prediction error metric \cite{XiangC_etal_2022_Accelerated_topology_optimization}. The results from this study highlight that this variability issue needs to be further addressed. So far, the authors have identified only a single study that investigated error variability when evaluating machine learning performance for civil and structural engineering applications \cite{TseranidisS_etal_2016_Data_driven_approximation_algorithms}. The use of the accuracy metric along with its minumum, maximum, 95\% and 99\% metrics could provide a framework to study error variability in more detail.

\subsection{Other neural network performance observations}
Despite using a simple multi-layer neural network for the structural design inverse operator $O^{\dagger}_{\mathrm{inv}}$, the development procedure successfully lowered the validation error from MAPE values of $\approx 10\%$ in Table \ref{table:results_50_50_loss_and_activation_comparison} to $1.6\%$ in Table \ref{table:results_final_model}, a performance that was matched by the testing dataset as well. This was attributable to numerous factors, a notable one being the use of the percentage based loss function, which as anticipated in Section \ref{sec:method_loss_functions}, was more suitable for the dataset given the orders of magnitude differences in the targets. The use of the exponential output activation function $a_{\mathrm{exp}}$ may also have positively contributed to dealing with target values that vary greatly in magnitude.

The lack of literature on machine learned structural design models for continuous beam systems means that a direct comparison of the $1.6\%$ MAPE performance is not possible at present. However, one can compare this performance with performance metrics of structural design models developed for different applications. For example, the network developed in this work outperformed previous multi-layer neural network regression models; an early concrete beam prediction model achieved a MAPE value of $10.17\%$ \cite{VanlucheneR_etal_1990_Neural_Networks_in}, whilst a cross-section predictor of aerospace components averaged out at a MAPE value of $5\%$ \cite{BerkeL_etal_1993_Optimum_design_of}. The network presented in this study also performed well when compared to more advanced network architectures such as convolutional neural networks for topologically optimised truss structures that achieved voxel value errors of $5.63\%$ \cite{XiangC_etal_2022_Accelerated_topology_optimization}. Comparison with further works that developed machine learned structural design models was not possible for studies which reported performance with non-percentage based metrics such as MAE \cite{KangH_etal_1994_Neural_Network_Approaches} or MSE \cite{TseranidisS_etal_2016_Data_driven_approximation_algorithms, BehzadiM_etal_2021_Real_Time_Topology_Optimization, YanJ_etal_2022_Deep_learning_driven}.

This study also differentiates itself by the quantity of data it was trained on (up to 700,000 data-points), which based on Figure \ref{fig:results_dataset_size_analysis} helped improve validation performance. Early works from the 1990s had training set sizes smaller than 100 data-points \cite{VanlucheneR_etal_1990_Neural_Networks_in, BerkeL_etal_1993_Optimum_design_of, KangH_etal_1994_Neural_Network_Approaches}, and even more recent literature only trained using 600 \cite{TseranidisS_etal_2016_Data_driven_approximation_algorithms}, 12,000 \cite{BehzadiM_etal_2021_Real_Time_Topology_Optimization} 28,000 \cite{XiangC_etal_2022_Accelerated_topology_optimization} or just under 40,000 \cite{NieZ_etal_2021_TopologyGAN__Topology_Optimization} data-points. Whilst large datasets significantly increase computational cost, the combination of big data and more advanced neural network architectures may improve performance further, both in terms of average error and prediction variability.

\subsection{Limitations and scope for future works}

There are multiple limitations that restrict practical use of the proposed design model. The first is the fact that the structural systems within the dataset were designed against ULS constraints only, and made other assumptions on the nature of the design problem listed in Section \ref{sec:problem_statement_detailed}. The generalisability of the model, specifically for system sizes $m<5$ also requires further work, and the issue of prediction variability will also require additional investigation in terms of either model architectures or generating larger datasets. Furthermore, there likely exist a wide range of mathematical techniques from inverse problems that could aid in developing and assessing operators for structural design problems, by for example estimating the Lipschitz coefficient of the mapping. These limitations provide a clear basis for further works in the future.

On another note, Table \ref{table:discussion_computation_time} summarises the total computation time required for the entirety of the results section. Whilst the computation time could have been accelerated through parallelisation, improved computation resources and simplification of metric evaluations algorithms, the purpose of Table \ref{table:discussion_computation_time} is to indicate the relative proportion of time spent at each stage. Greater computational resources may allow investigations using alternative validation strategies such as \textit{k}-fold cross-validation \cite{KimJ_2009_Estimating_classification_error} and automated hyperparameter selection procedures \cite{VakhariaV_etal_2023_Estimation_of_Lithium_ion} that could result in improved performance. A significant portion of the computation effort was spent simply generating the data-points for training, validation and testing. In light of encouraging reproducibility studies \cite{SunH_etal_2021_Machine_learning_applications} and to encourage research that improves the predictive capability of the machine learned structural design model presented here, the CBeamXP dataset along with an associated python-based neural network training script are made available at an open-source data repository \cite{GalletA_etal_2023_CBeamXP__Continuous_Beam}.

\begin{table}[htb]
    \centering
    \begin{tabular}{rlccc}
        \toprule
        
        \multirow[c]{2}{2cm}[-0.10cm]{\centering Section} & \multirow[c]{2}{4cm}[-0.10cm]{\centering Stage} & \multicolumn{2}{c}{\centering Computation time} & \multirow[c]{2}{2cm}[-0.10cm]{\centering Proportion [\%]} \\ \cmidrule{3-4} 
        
        & & In hours [h] & In days [D] & \\ \midrule
        
        Section \ref{sec:results_influence_zone}  & Influence zone evaluation & 5 & 0.2 & \multirow[c]{2}{2cm}[-0.10cm]{\centering 33} \\
        
        Section \ref{sec:results_data_generation_prepocessing} & Data generation and pre-processing & 84 & 3.5 & \\ \midrule
        
        Section \ref{sec:results_loss_and_activation} & Loss and activation function study & 5 & 0.2 & \multirow[c]{3}{2cm}[-0.10cm]{\centering 27} \\
        
        Section \ref{sec:results_height_and_depth} & Height and depth study & 34 & 1.4 & \\
        
        Section \ref{sec:results_dataset_size} & Dataset size study & 36 & 1.5 & \\ \midrule
        
        Section \ref{sec:results_final_model_and_robustness} & Testing and robustness study & 108 & 4.5 & 40 \\ \midrule
        
        \multicolumn{2}{c}{\textbf{Total computation time:}} & 272 & 11.3 & 100 \\
        
        \bottomrule
    \end{tabular}
    \caption{Computation time for each neural network development stage.}
    \label{table:discussion_computation_time}
\end{table}

%% file: 06_Conclusion.tex
\section{Conclusions}
\label{sec:conclusion}

This work developed a new neural network based structural design model to predict cross-section property requirements of continuous beam systems non-iteratively. The major contributions of this investigation include:

\begin{itemize}
    \item Framing structural design as an inverse problem, and using this novel perspective to identify three distinct types of machine learning applications. One of these types, machine learned inverse operators, were investigated in this work to develop a non-iterative structural design model. This presents a fundamental shift from traditional design approaches.
    \item Developing a non-iterative structural design model for continuous beam systems of arbitrary member size through the novel use of influence zones \cite{GalletA_etal_2023_Influence_zones_for} to provide a mechanics-driven feature selection process that enhanced the model's generalisability.
    \item Achieving a mean absolute percentage error of 1.6\% which was lower than machine learned structural design models from comparative literature. This performance was attributable to the careful consideration of the network architecture in terms of height and depth of the hidden layers, the selection of loss and activation functions that were appropriate to address the challenges posed by continuous beam system, and a dataset size of 700,000 data points.
    \item Identifying the importance of measuring and reducing prediction error variability. In this study the 99\% confidence interval for testing accuracy was between 91.7\% and 113.8\%. Reducing prediction variability is a significant knowledge gap in literature, especially in regards to machine learning applications within safety critical systems such as structural design.
\end{itemize}

The CBeamXP dataset generated in this work containing one million data-points along with an associated python-based neural network training script were published at an open-source data repository \cite{GalletA_etal_2023_CBeamXP__Continuous_Beam}. Aside from allowing results to be reproduced, sharing this data will hopefully encourage future research towards machine learned structural design models that improve the mean absolute percentage error, generalisability, or prediction variability achieved in this investigation.

%% file: 07_Acknowledgements.tex
\subsection*{Declaration of conflicting interests}
\noindent The author(s) declared no potential conflicts of interest with respect to the research, authorship, and/or publication of this article.

\subsection*{Data statement}
\noindent The dataset generated and used in this work as well as a neural network training script are available at an open-source data repository \cite{GalletA_etal_2023_CBeamXP__Continuous_Beam}.

\subsection*{Funding}
\noindent The author(s) disclosed receipt of the following financial support for the research, authorship, and/or publication of this article. The authors gratefully acknowledge the support of the Engineering Physical Sciences Research Council's (EPSRC) Doctoral Training Partnership Studentship and the Ramboll Foundation. For the purpose of open access and funding stipulations, the authors have applied a Creative Commons Attribution (CC-BY) licence to any Accepted Manuscript (AM) version arising.

\subsection*{CRediT authorship contribution statement}
\noindent \textbf{Adrien Gallet}: Conceptualisation, Methodology, Software, Investigation, Formal analysis, Validation, Data curation, Visualisation, Writing - Original Draft, Writing - Review \& Editing, Project administration. \textbf{Andrew Liew}: Writing - Review \& Editing. \textbf{Iman Hajirasouliha}: Validation, Writing - Review \& Editing. \textbf{Danny Smy}l: Supervision, Validation, Writing - Review \& Editing.